\documentclass{article}

\usepackage{arxiv}
\usepackage[utf8]{inputenc}
\usepackage[T1]{fontenc}
\usepackage{hyperref}
\usepackage{url}
\usepackage{booktabs}
\usepackage{amsfonts}
\usepackage{nicefrac}
\usepackage{microtype}
\usepackage{graphicx}
\usepackage{algorithm}
\usepackage{algorithmic}
\usepackage{amsmath}
\usepackage{enumitem}
\usepackage{tabularx}
\usepackage{multirow}
\usepackage{float}
\usepackage{natbib}
\usepackage{xcolor}
\usepackage{colortbl}

\title{FAB-Bench: A Framework for Adaptive RAG Benchmarking in Semiconductor Manufacturing}

\author{
  Jingbin Qian \quad
  Congwen Yi \quad
  Min Xia \quad
  Wen Wu \quad
  Jun Zhu \quad
  Jian Guan\textsuperscript{*}\\[6pt]
  FutureFab.AI \\[2pt]
  \textsuperscript{*} andrewg@futurefab.ai
}

\date{}

\newcommand\blfootnote[1]{%
  \begingroup
  \renewcommand\thefootnote{}\footnote{#1}%
  \addtocounter{footnote}{-1}%
  \endgroup
}

\usepackage{fancyhdr}
\makeatletter
\fancypagestyle{headings}{%
  \fancyhf{}%
  \fancyhead[LO]{\fontsize{9}{11}\selectfont\bfseries FAB-Bench}%
  \fancyhead[RE]{\fontsize{9}{11}\selectfont\bfseries FAB-Bench}%
  \fancyhead[RO,LE]{\fontsize{9}{11}\selectfont\bfseries A {P\scriptsize{REPRINT}}}%
}
\makeatother

\begin{document}

\maketitle
\blfootnote{Benchmark dataset available at: \url{https://github.com/FuturefabAI/FAB-Bench}}
\begin{abstract}
Retrieval-Augmented Generation (RAG) has become critical for knowledge-intensive applications, yet evaluating its performance in vertical domains remains difficult due to domain complexity, diverse context scales, and heavy reliance on expert assessments that are costly, inconsistent, and non-scalable.
We introduce \textbf{FAB-Bench}, an end-to-end \textbf{f}ramework for \textbf{a}daptive \textbf{b}enchmarking of RAG systems in semiconductor manufacturing. FAB-Bench defines six diagnostic metrics measuring factual accuracy, contextual utilization, completeness, retrieval relevance, technical depth, and reasoning consistency. The framework couples retriever diagnostics with generator-level reasoning analysis across context windows of 4K--32K tokens, quantifying how retrieval precision and generative fidelity co-evolve as contextual scope expands. From over 1,300 generated candidates, we curated a high-quality benchmark of 200 query--answer pairs spanning three synthesis strategies: needle-in-haystack, intra-document multi-topic, and cross-document multi-hop. Systematic evaluation across four LLMs and four RAG frameworks reveals three distinct context-scaling behaviors---logarithmic growth, early saturation, and cold-start dynamics---and identifies attention dilution as the primary mechanism behind performance degradation at extreme context lengths. Cross-framework validation on three additional production RAG systems confirms evaluation portability.
\end{abstract}


\section{Introduction}
\label{sec:intro}

Large language models (LLMs) have demonstrated remarkable capabilities in various
tasks~\citep{achiam2023gpt,brown2020gpt3}, motivating the development of diverse evaluation benchmarks.
Early benchmarks such as GLUE~\citep{wang2018glue} and SuperGLUE~\citep{wang2019superglue}
focused on natural language understanding tasks, including sentiment analysis, textual understanding, and question answering.
\textbf{MMLU}~\citep{hendrycks2021mmlu} expands assessment to broad knowledge coverage across 57 subjects using 15,908 multiple-choice questions spanning from elementary to professional levels, with an emphasis on zero-shot and few-shot evaluation of pre-trained knowledge. However, rapid benchmark saturation has significantly reduced its discriminative power: while GPT-3 achieved
only 43.9\% accuracy~\citep{brown2020gpt3}, GPT-4 exceeded
86\%~\citep{openai2023gpt4}. To improve linguistic coverage, \textbf{C-Eval}~\citep{huang2023ceval} extends this paradigm to Chinese with 13,948 questions across 52 subjects. Despite their value, these benchmarks share fundamental limitations: reliance on static \textit{public} knowledge and the absence of retrieval contexts, making them poorly suited for evaluating RAG systems in vertical domains.

Beyond generic language benchmarks, more task- or capability-specific evaluations have been proposed.
\textbf{ARC}~\citep{clark2018arc} evaluates scientific reasoning using graduate-school--level questions;
\textbf{TruthfulQA}~\citep{lin2022truthfulqa} measures propensity for factual reliability; \textbf{GSM8K}~\citep{cobbe2021gsm8k}
and \textbf{MATH}~\citep{hendrycks2021math} assess mathematical reasoning from
graduate-school to competition-level; and \textbf{HumanEval}~\citep{chen2021humaneval}
evaluates code generation through 164 programming tasks.

Domain-specific benchmarks further address specialized requirements. In healthcare, \textbf{MedQA}~\citep{jin2021medqa}, \textbf{MedMCQA}~\citep{pal2022medmcqa}, and \textbf{MultiMedQA}~\citep{singhal2023medpalm} evaluate medical reasoning under safety-critical constraints, with Med-PaLM 2 achieving 85\% accuracy on USMLE-style questions~\citep{singhal2023medpalm2}. Legal benchmarks such as \textbf{LegalBench}~\citep{guha2023legalbench} and \textbf{LawBench}~\citep{fei2023lawbench} assess regulatory reasoning across jurisdictions, while financial benchmarks including \textbf{FinanceBench}~\citep{financebench}, \textbf{FinBen}~\citep{xie2024finben}, and \textbf{BloombergGPT}~\citep{wu2023bloomberggpt} evaluate financial analysis and decision-making. In semiconductor design, ChipNeMo~\citep{liu2023chipnemo} adapts LLMs for chip design but focuses on model training rather than RAG evaluation. Although these domain benchmarks improve specialization, they largely rely on public datasets and expert curation, and provide limited visibility into how effectively RAG systems retrieve, integrate, and reason over proprietary, multi-document corpora.

RAG has become the dominant paradigm for deploying LLMs in knowledge-intensive applications, particularly in enterprise and industrial settings~\citep{lewis2020retrieval}---where models must reason and answer questions over proprietary documents not seen during training. Despite the progress of domain-specific benchmarks, many remain poorly suited for enterprise RAG use cases: they typically evaluate over \textit{public} knowledge sources (e.g., medical licensing exams, legal case law, published financial reports) rather than the proprietary documentation enterprises actually deploy~\citep{chen2024rethinking}. Moreover, heavy dependence on manual expert curation fundamentally limits scalability~\citep{zheng2024judging}. Existing evaluation metrics are often accuracy-based, offering little diagnostic insight into whether failures arise from incomplete retrieval, faulty reasoning, or inadequate multi-document synthesis. They also lack mechanisms for systematic deployment optimization: they cannot evaluate RAG effectiveness on proprietary corpora, quantify knowledge augmentation value, or guide configuration decisions such as context window allocation. Thus, these benchmarks emphasize knowledge recall or task-specific reasoning while providing limited assessment of workflow-level integration or nuanced judgment in real-world settings~\citep{budler2025healthcare,gao2023rag}. While recent work has begun to address RAG-specific evaluation, important gaps remain: RGB~\citep{chen2024benchmarking} and RECALL~\citep{liu2023recall} focus on general-domain QA rather than specialized knowledge, ARES~\citep{saad2024ares} requires substantial human calibration, and RAGAS~\citep{es2024ragas} lacks domain-specific customization. As a result, enterprises still lack quantitative guidance for deployment-critical decisions such as model selection and context window allocation, relying instead on ad-hoc qualitative feedback.

\subsection{Contributions}
In this work, we introduce \textbf{FAB-Bench}, an end-to-end evaluation methodology for vertical-domain RAG for realistic enterprise reasoning. Our main contributions are summarized as follows:
\begin{itemize}
    \item \textbf{Methodology for evaluating vertical-domain RAG via cross-document synthesis under adaptive benchmark generation.}
    We formulate vertical-domain RAG evaluation as evidence-based synthesis over long and heterogeneous private corpora, and design benchmarks that require explicit multi-document integration, including needle-in-haystack grounding, intra-document multi-topic reasoning, and cross-document multi-hop composition. To improve robustness and coverage of synthesized queries, we employ an adaptive generation mechanism with temperature modification, adjusting sampling temperature in response to quality and consistency signals to obtain diverse and stable benchmark instances.

    \item \textbf{A diagnostic measurement protocol that attributes failures across retrieval and generation.}
    We introduce a six-dimensional evaluation rubric---Completeness, Factuality, Context Utilization, Technical Depth, Relevance, and Support Quality---that separates missing evidence, irrelevant retrieval, shallow synthesis, and unsupported generation, enabling fine-grained localization of performance bottlenecks.

    \item \textbf{An empirical characterization of context-window scaling regimes for vertical RAG.}
    By measuring performance from 4K to 32K tokens across four LLMs, we identify three distinct scaling behaviors and characterize attention dilution through metric-level decomposition, offering actionable guidance for configuration decisions.
\end{itemize}

\section{Related Work}
\label{sec:related_work}

\subsection{LLM Evaluation Benchmarks}

General-purpose benchmarks have evolved from task-specific assessments like GLUE~\citep{wang2018glue} and SuperGLUE~\citep{wang2019superglue} to broad knowledge evaluations. MMLU~\citep{hendrycks2021mmlu} covers 57 subjects with 15,908 questions, though rapid saturation (GPT-3: 43.9\%~\citep{brown2020gpt3} to GPT-4: 86\%+~\citep{openai2023gpt4}) reduces its discriminative power. C-Eval~\citep{huang2023ceval} extends coverage to Chinese. Capability-specific benchmarks target scientific reasoning (ARC~\citep{clark2018arc}), factual reliability (TruthfulQA~\citep{lin2022truthfulqa}), mathematical reasoning (GSM8K~\citep{cobbe2021gsm8k}, MATH~\citep{hendrycks2021math}), and code generation (HumanEval~\citep{chen2021humaneval}). These benchmarks share a fundamental limitation: reliance on static, public knowledge without retrieval contexts, making them poorly suited for RAG evaluation.

\subsection{Domain-Specific Evaluation}

Domain benchmarks address specialized requirements but inherit similar limitations. In healthcare, MedQA~\citep{jin2021medqa}, MedMCQA~\citep{pal2022medmcqa}, and MultiMedQA~\citep{singhal2023medpalm} evaluate medical reasoning, with Med-PaLM 2 reaching 85\% on USMLE-style questions~\citep{singhal2023medpalm2}. Legal benchmarks (LegalBench~\citep{guha2023legalbench}, LawBench~\citep{fei2023lawbench}) assess regulatory reasoning, while financial benchmarks (FinanceBench~\citep{financebench}, FinBen~\citep{xie2024finben}, BloombergGPT~\citep{wu2023bloomberggpt}) evaluate financial analysis. In semiconductor design, ChipNeMo~\citep{liu2023chipnemo} adapts LLMs for chip design but focuses on model training rather than RAG evaluation. These benchmarks primarily evaluate parametric knowledge over public datasets, providing limited visibility into how RAG systems retrieve, integrate, and reason over proprietary multi-document corpora~\citep{budler2025healthcare}.

\subsection{RAG Evaluation Frameworks}

RAG-specific evaluation has received increasing attention. RAGAS~\citep{es2024ragas} provides multi-dimensional metrics (faithfulness, relevance, context precision/recall) and supports test generation from user-provided corpora, but does not address context-window scaling or domain-specific metric customization. ARES~\citep{saad2024ares} automates RAG evaluation through prediction-powered inference with multi-dimensional scoring, but requires $\sim$150 human-annotated samples for calibration. RGB~\citep{chen2024benchmarking} evaluates four RAG robustness abilities including information integration across documents, but uses a fixed general-domain dataset. RECALL~\citep{liu2023recall} evaluates robustness against counterfactual knowledge. CRAG~\citep{yang2024crag} provides a comprehensive benchmark with multi-hop and aggregation questions requiring cross-document synthesis, but operates on a fixed dataset without vertical-domain customization. MultiHop-RAG~\citep{tang2024multihoprag} specifically targets multi-hop reasoning with evidence distributed across 2--4 documents, but is limited to a fixed English news corpus. SCARF~\citep{rengo2025system} proposes a system-level assessment framework but does not include benchmark generation.

FAB-Bench complements these efforts by addressing two gaps that none of the above frameworks cover simultaneously: (1)~\textit{systematic context-window scaling analysis} that characterizes how RAG performance evolves from 4K to 32K tokens, and (2)~\textit{domain-specific evaluation with a structured knowledge base} (431 semiconductor terms across 7 weighted categories) that enables precision-aware benchmark generation and domain-grounded scoring.

\subsection{LLM-as-Judge Methodology}

Using LLMs as evaluation judges has become widespread following MT-Bench~\citep{zheng2024judging}, which demonstrated strong correlation between LLM judgments and human preferences. G-Eval~\citep{liu-etal-2023-g} formalizes this through chain-of-thought prompting with probability-weighted scoring. However, LLM judges exhibit known biases including position bias, verbosity bias, and self-enhancement bias~\citep{zheng2024judging,wang2023large}. Recent work on calibrating LLM judges~\citep{liu2024aligning} suggests that structured rubrics with explicit scoring criteria mitigate these biases. Our approach addresses reliability through: (1)~structured rubrics with separate objective/subjective variants per metric; (2)~chain-of-thought reasoning via G-Eval; and (3)~empirical validation of metric independence (Section~\ref{sec:metric_independence}).

\section{FAB-Bench Framework}
\label{sec:framework}

\begin{figure}[htbp]
    \centering
    \includegraphics[width=0.8\linewidth]{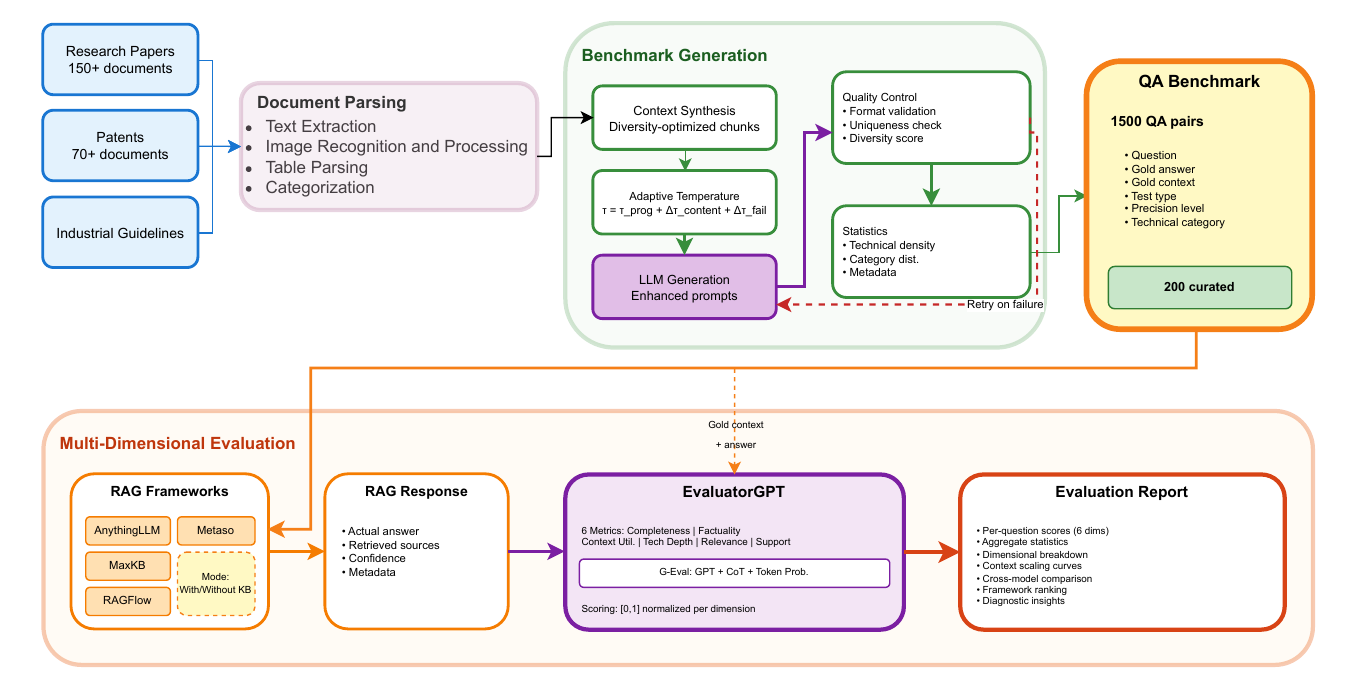}
    \caption{FAB-Bench system overview. The framework comprises two components: an adaptive benchmark generation system that produces domain-specific QA pairs from proprietary corpora, and a multi-dimensional evaluation platform that scores RAG system responses across six diagnostic metrics.}
    \label{fig:system_overview}
\end{figure}

Evaluating RAG systems in vertical domains requires benchmarks satisfying four criteria: \textit{authenticity} (questions reflecting real-world complexity), \textit{contamination resistance} (avoiding train-test overlap), \textit{diagnostic granularity} (isolating specific failure modes), and \textit{discriminative power} (meaningfully differentiating models). FAB-Bench addresses these through two design principles: evaluation mirrors deployment by generating questions from user-provided corpora, and multi-dimensional assessment enables precise failure attribution (Figure~\ref{fig:system_overview}).

\subsection{Knowledge Corpus and Domain Encoding}
\label{sec:corpus}

Our benchmark draws from three source types: academic literature comprising 150+ peer-reviewed papers from IEDM, ISSCC, and VLSI symposia; patent documents with 70+ filings containing proprietary fabrication details; and industry standards from SEMI specifications. The corpus totals approximately 347 million tokens across 188 topics.

To enable domain-aware processing, we constructed a hierarchical knowledge base $\mathcal{K}$ encoding 431 technical terms organized into seven semantic categories (Table~\ref{tab:knowledge_base}), each with precision weights $w_i$ reflecting technical rigor requirements. The knowledge base serves three functions: (1)~computing technical density $\rho(d)$ for adaptive generation control, (2)~classifying document precision levels for temperature scheduling, and (3)~weighting domain terminology in similarity computations.

\begin{table}[H]
\centering
\small
\caption{Hierarchical knowledge base: seven semantic categories with term counts and precision weights. Higher weights indicate categories requiring greater quantitative precision.}
\label{tab:knowledge_base}
\begin{tabular}{llcl}
\toprule
\textbf{Category} & \textbf{Example Terms} & \textbf{Terms} & \textbf{Weight} \\
\midrule
Performance Parameters & $V_{th}$, leakage current, $f_T$, DIBL & 80 & 1.7 \\
Process Nodes & EUV, FinFET, HKMG, SADP, TSV & 65 & 1.6 \\
Manufacturing Processes & ALD, CMP, RIE, PECVD, epitaxy & 60 & 1.5 \\
Device Physics & GAAFET, MRAM, nanosheet, CFET & 62 & 1.4 \\
Materials Science & HfO$_2$, low-$k$, SiGe, ruthenium & 51 & 1.3 \\
Testing Methodologies & NBTI, TDDB, CD-SEM, electromigration & 51 & 1.2 \\
Applications & AI chip, NPU, ADAS, 5G transceiver & 62 & 1.1 \\
\midrule
\textbf{Total} & & \textbf{431} & \\
\bottomrule
\end{tabular}
\end{table}

Questions span two balanced formats. Objective questions (50\%) assess factual accuracy through mathematical calculations, fill-in-blank, true/false, and multiple choice items. Subjective questions (50\%) evaluate reasoning through mechanism explanations, causal reasoning, comparative analysis, and problem diagnosis. Both formats require context-dependent answers that cannot be resolved from parametric knowledge alone.

\subsection{Cross-Document Context Synthesis}
\label{sec:context_synthesis}

A defining RAG capability is cross-document synthesis---integrating information across multiple sources. Single-document benchmarks cannot distinguish retrieval-based reasoning from parametric knowledge reliance. We design three synthesis strategies targeting complementary objectives.

The corpus is segmented into chunks $C = \bigcup_{i=1}^{m} \text{Chunk}(D_i)$ via sliding window (512 tokens, 128-token overlap). TF-IDF vectors $\vec{v}(c)$ enable semantic comparison between chunks.

\paragraph{Strategy 1: Needle-in-Haystack.}
Critical information is embedded within topically dissimilar distractors. Given a target chunk $c_{\text{target}}$, we select distractor chunks $\{c_d\}$ that minimize cosine similarity: $c_d = \arg\min_{c \in C} \cos(\vec{v}(c_{\text{target}}), \vec{v}(c))$. This tests precise fact location amid irrelevant content.

\paragraph{Strategy 2: Intra-Document Multi-Topic.}
Chunks from different topic clusters within the same document are combined, requiring integration of dispersed information. We cluster chunks within each document using TF-IDF cosine similarity and select chunks from at least two distinct clusters ($\cos(\vec{v}(c_i), \vec{v}(c_j)) < 0.3$ for cluster separation).

\paragraph{Strategy 3: Cross-Document Multi-Hop.}
The most challenging strategy constructs contexts requiring cross-source reasoning:
\begin{enumerate}[leftmargin=*, itemsep=1pt]
    \item Select a seed chunk $c_{\text{seed}}$ from document $D_i$.
    \item Identify a semantically related chunk from a different document: $c_{\text{link}} = \arg\max_{c \in D_j, j \neq i} \cos(\vec{v}(c_{\text{seed}}), \vec{v}(c))$.
    \item Validate connection strength: $\cos(\vec{v}(c_{\text{seed}}), \vec{v}(c_{\text{link}})) > \theta_{\text{link}}$, where $\theta_{\text{link}} = 0.1$.
    \item Construct the final context by combining seed, link, and distractor chunks.
\end{enumerate}
Questions are generated \textit{after} context construction, guaranteeing that correct answers depend on synthesizing linked information from multiple documents.

\subsection{Adaptive Generation Control}
\label{sec:generation_control}

Generating high-quality test cases requires balancing precision and diversity. We adapt generation parameters to content characteristics through two mechanisms.

\paragraph{Technical Density and Precision Classification.}
We compute technical density $\rho(d)$ as the ratio of domain term occurrences (weighted by category) to total words. Documents are classified into precision levels:
\begin{equation}
p(d) =
\begin{cases}
\text{high} & \text{if } \rho(d) > 0.20 \text{ or } \omega_h(d) > 8 \\
\text{medium} & \text{if } \rho(d) > 0.12 \text{ or } \omega_h(d) > 4 \\
\text{low} & \text{otherwise}
\end{cases}
\label{eq:precision}
\end{equation}
where $\omega_h(d) = \sum_{c_i \in \{\text{parameters, processes}\}} w_i \cdot |\{t \in T_{c_i} : n(t, d) > 0\}|$ counts high-weight category term occurrences.

\paragraph{Adaptive Temperature.}
Temperature $\tau$ is computed as:
\begin{equation}
\tau = \text{clip}\left(\tau_{\min}(p) + \tau_{\text{prog}}(k) + \Delta\tau_c(c^*) + \Delta\tau_{\text{fail}}(a, s),\; 0.1,\; 1.0\right)
\label{eq:temperature}
\end{equation}
where base ranges $[\tau_{\min}, \tau_{\max}]$ are $[0.4, 0.8]$ for high-precision, $[0.5, 0.9]$ for medium, and $[0.6, 1.0]$ for low; $\tau_{\text{prog}}(k) = (\tau_{\max} - \tau_{\min}) \times \min(k/20, 0.8)$ increases diversity as successful generations $k$ accumulate; $\Delta\tau_c$ adjusts for dominant content category ($-0.10$ for parameters, $+0.05$ for applications); and $\Delta\tau_{\text{fail}}$ boosts temperature after generation failures. These mechanisms follow established temperature-diversity trade-offs~\citep{Holtzman2020The,renze2024effect}.

\paragraph{Adaptive Similarity Threshold.}
To prevent duplicate generation while allowing necessary terminology overlap in technical content, the similarity threshold relaxes progressively:
\begin{equation}
\theta_{\text{sim}}(r) = \max\left(0.50,\; \theta_{\text{base}}(p) - 0.05 \times r\right)
\label{eq:threshold}
\end{equation}
where $\theta_{\text{base}}$ is 0.70 for high-precision, 0.75 for medium, and 0.80 for low-precision content, and $r$ counts retry attempts. Similarity combines weighted Jaccard overlap of all tokens and domain terms with TF-IDF cosine similarity for corpora exceeding five questions.

\paragraph{Generation Pipeline.}

\begin{figure}[htbp]
    \centering
    \includegraphics[width=0.5\linewidth]{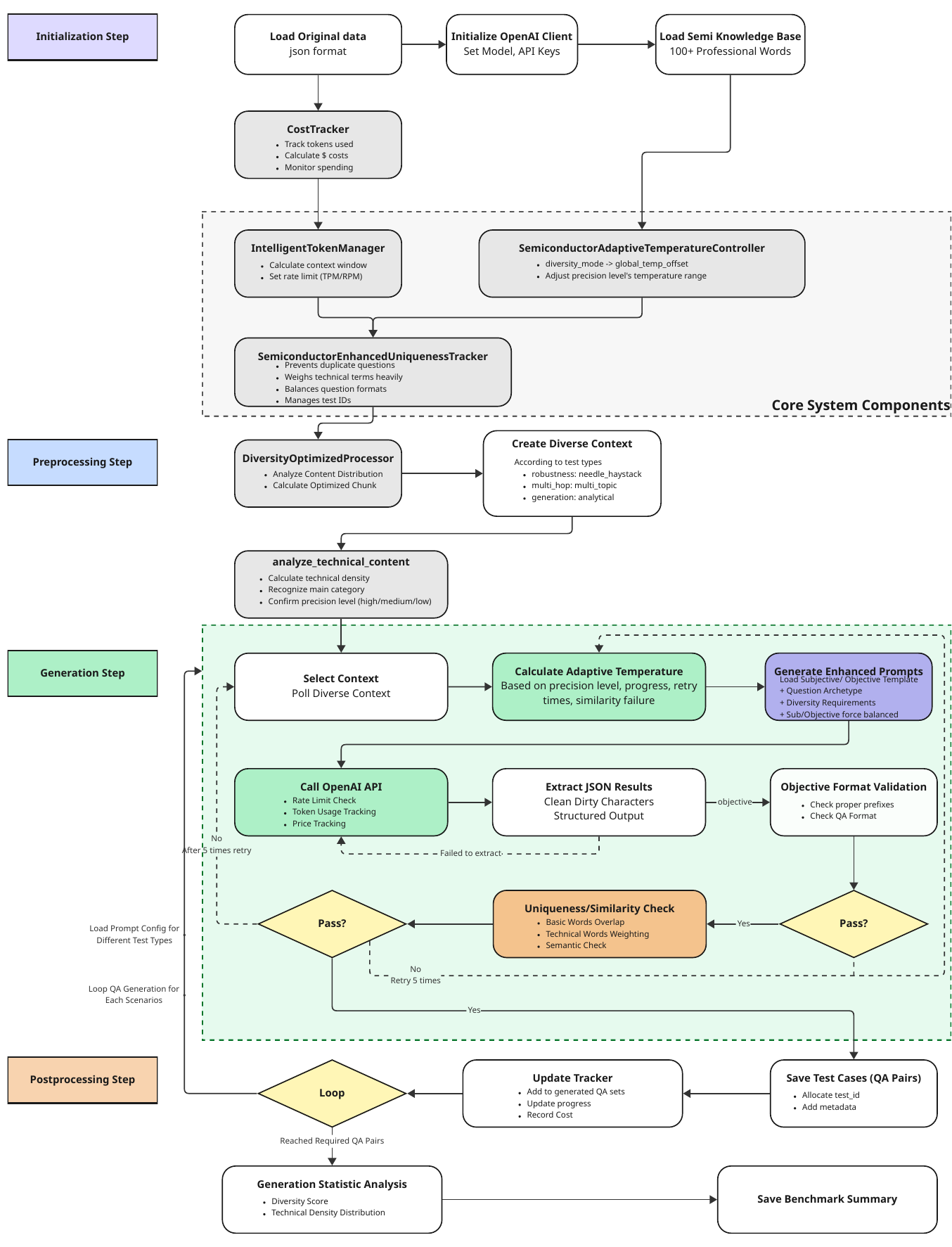}
    \caption{Benchmark generation workflow. The pipeline iterates through synthesized contexts, adaptively adjusting generation parameters based on precision classification and failure feedback.}
    \label{fig:workflow_embedded}
\end{figure}

Algorithm~\ref{alg:question_generation} integrates these components into a fully automated pipeline (Figure~\ref{fig:workflow_embedded}).

\begin{algorithm}[H]
\caption{Adaptive QA Pair Generation}
\label{alg:question_generation}
\small
\begin{algorithmic}[1]
\REQUIRE Corpus $D$, test type $t$, target count $N$
\ENSURE QA pairs $\mathcal{Q} = \{(q_i, d_i, a_i)\}_{i=1}^{N}$
\STATE $C \gets \textsc{Chunk}(D)$; compute TF-IDF vectors $\vec{v}(c)$ for all $c \in C$
\STATE $\mathcal{D} \gets \textsc{SynthesizeContexts}(C, t)$ \hfill\textit{// Strategy 1, 2, or 3}
\STATE $\mathcal{Q} \gets \emptyset$
\FOR{each context $d \in \mathcal{D}$ \textbf{while} $|\mathcal{Q}| < N$}
    \STATE Classify precision $p(d)$ via Eq.~\ref{eq:precision}; compute density $\rho(d)$
    \STATE $\tau \gets \textsc{AdaptiveTemperature}(p, |\mathcal{Q}|, \text{failures})$ \hfill\textit{// Eq.~\ref{eq:temperature}}
    \STATE $\theta \gets \textsc{AdaptiveThreshold}(p, \text{retries})$ \hfill\textit{// Eq.~\ref{eq:threshold}}
    \FOR{retry $r = 1$ to $R_{\max}$}
        \STATE $(q, a) \gets \textsc{LLMGenerate}(d, t, \tau)$
        \IF{\textsc{Valid}$(q, a)$ \AND \textsc{Unique}$(q, \mathcal{Q}, \theta)$}
            \STATE $\mathcal{Q} \gets \mathcal{Q} \cup \{(q, d, a)\}$; \textbf{break}
        \ENDIF
    \ENDFOR
\ENDFOR
\RETURN $\mathcal{Q}$
\end{algorithmic}
\end{algorithm}

\paragraph{Expert Validation.}
A controlled ablation study validated the generation mechanism. Domain experts rated QA pairs (18 per condition, spanning ROB/MULTI/GEN types equally) on four dimensions (1--5 scale): accuracy, relevance, difficulty, and diversity. The full system combining adaptive parameters with enhanced prompts achieved the highest scores across all dimensions (Accuracy: 5.00, Relevance: 5.00, Difficulty: 4.60, Diversity: 4.53) with zero retries, while adaptive parameters alone yielded mixed results (Accuracy: 4.33, Difficulty: 3.28), indicating that prompt engineering is essential for quality generation (Appendix~\ref{app:expert_evaluation}).

Per-question-type analysis reveals an appropriate difficulty gradient in the generated benchmark (Table~\ref{tab:question_type_quality}), confirming that the three synthesis strategies produce questions of distinct and meaningful complexity.

\begin{table}[H]
\centering
\small
\caption{Expert-rated quality by question type (averaged across all ablation conditions). The difficulty gradient (ROB $<$ MULTI $<$ GEN) confirms the three synthesis strategies produce appropriately graded complexity.}
\label{tab:question_type_quality}
\begin{tabular}{lcccc}
\toprule
\textbf{Question Type} & \textbf{Accuracy} & \textbf{Relevance} & \textbf{Difficulty} & \textbf{Diversity} \\
\midrule
Robustness (needle-in-haystack) & 4.58 & 4.62 & 2.38 & 2.92 \\
Multi-Hop (cross-document) & 4.58 & 4.83 & 4.12 & 3.33 \\
Generation (deep reasoning) & 4.71 & 4.75 & 4.79 & 3.88 \\
\bottomrule
\end{tabular}
\end{table}

\subsection{Six-Dimensional Evaluation Metrics}
\label{sec:metrics}

We define six diagnostic metrics, each evaluated on a 10-point scale with separate rubrics for objective and subjective questions (Table~\ref{tab:metrics}). Full rubric definitions are provided in Appendix~\ref{app:rubrics}.

\begin{table}[H]
\centering
\small
\caption{Six-dimensional evaluation metrics with failure mode attribution.}
\label{tab:metrics}
\begin{tabular}{@{}lll@{}}
\toprule
\textbf{Metric} & \textbf{Evaluates} & \textbf{Failure Attribution} \\
\midrule
Completeness & Coverage of required elements & Missing retrieval or shallow generation \\
Technical Depth & Sophistication of analysis & Generation: insufficient reasoning \\
Factuality & Alignment with sources & Generation: hallucination \\
Relevance & Focus on query topic & Both: off-topic retrieval or generation \\
Context Utilization & Use of retrieved context & Retrieval failure or context neglect \\
Support Quality & Citation specificity/accuracy & Generation: unsupported claims \\
\bottomrule
\end{tabular}
\end{table}

These metrics are designed to capture orthogonal failure modes. A response may exhibit high relevance but low factuality (on-topic hallucination), or strong technical depth but poor context utilization (ignoring retrieved documents in favor of parametric knowledge). This design enables diagnostic attribution: retrieval failures manifest as low context utilization despite high factuality, while generation failures appear as uniformly low scores.

We empirically assess metric independence in Section~\ref{sec:metric_independence} through correlation analysis across all experimental configurations.

\subsection{Evaluation Platform Architecture}
\label{sec:platform_architecture}

Our platform implements a three-layer architecture inspired by SCARF~\citep{rengo2025system}:

\begin{itemize}[leftmargin=*, itemsep=2pt]
    \item \textbf{Orchestration Layer:} Manages test distribution, parallel execution across models and context configurations, and result aggregation.
    \item \textbf{Adapter Layer:} Normalizes heterogeneous RAG framework interfaces (AnythingLLM, RAGFlow, MaxKB, Metaso) into unified query-response protocols, enabling fair cross-framework comparison. For frameworks without source exposure (e.g., MaxKB), the adapter employs heuristic context detection using domain-specific indicators.
    \item \textbf{Evaluation Layer:} Applies the six-dimensional metrics via G-Eval~\citep{liu-etal-2023-g} through DeepEval~\citep{deepeval}. G-Eval employs LLMs as judges with chain-of-thought reasoning: each metric is defined through evaluation objectives, reasoning procedures, and scoring rubrics. The evaluator extracts token-level probabilities over score candidates and computes normalized weighted scores.
\end{itemize}

The platform supports two complementary evaluation paradigms: \textit{cross-model comparison} (multiple LLMs within a fixed RAG framework, isolating model capabilities) and \textit{cross-framework comparison} (multiple RAG architectures with a fixed model, isolating retrieval and system design choices).

\section{Experiments: Context Window Scaling Analysis}
\label{sec:experiment}

\begin{figure}[H]
    \centering
    \includegraphics[width=\textwidth]{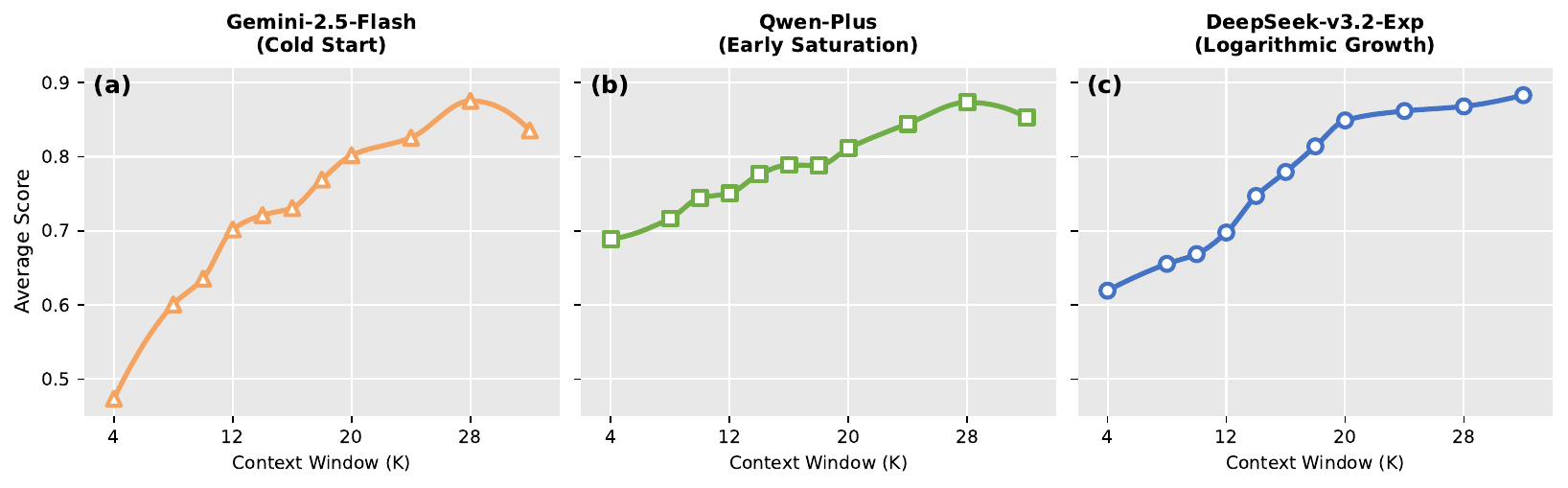}
    \caption{Aggregate performance trajectories across context windows (4K--32K). Three distinct scaling behaviors emerge: logarithmic growth (DeepSeek), early saturation (Qwen-Plus), and cold-start dynamics (Gemini).}
    \label{fig:scaling_overall}
\end{figure}

\subsection{Experimental Configuration}

We conducted context-scaling analysis across four LLMs: DeepSeek-v3.2-Exp (2025-09-29), Qwen-Plus (2025-09-11), Gemini-2.5-Flash, and Qwen-2.5-72B-Instruct. All experiments used AnythingLLM as the unified RAG framework with a fixed, pre-processed segmented JSON corpus to minimize confounding variables.

\paragraph{Context Window and Output Settings.}
Context window size was varied via AnythingLLM's workspace configuration through OpenAI-compatible APIs. Output token limits scaled with context: 1K for 4K context, 2K for 8K--10K, and 4K for $\geq$12K. We evaluated 11 configurations from (4K, 1K) to (32K, 4K), with finer granularity in the 10K--20K range where preliminary results indicated performance transitions. For Qwen-2.5-72B, we additionally tested extended contexts at 64K and 128K tokens.

\paragraph{Evaluation Protocol.}
Each model--context configuration was evaluated on the full 200-question benchmark. The benchmark comprises 59 robustness questions (needle-in-haystack), 90 multi-hop reasoning questions (cross-document synthesis), and 51 generation quality questions (intra-document multi-topic). Responses were scored using GPT-4.1-mini via DeepEval's G-Eval implementation.

\paragraph{Statistical Scale.}
The evaluation encompasses 200 questions $\times$ 4 models $\times$ 11+ configurations $\times$ 6 metrics = over 52,800 individual metric evaluations, providing sufficient statistical power for the reported comparisons.

\subsection{Overall Scaling Trends}

Figure~\ref{fig:scaling_overall} presents aggregate performance trajectories for the three primary models.

\paragraph{Convergent Performance at Scale.}
At 4K context, model differences are pronounced: Gemini achieves only 0.474, while Qwen-Plus leads at 0.689 and DeepSeek reaches 0.619---a 45\% gap between best and worst. This spread narrows substantially as context increases: by 20K, all models cluster between 0.80--0.85, and at 28K, performance converges within a 1\% band (0.868--0.876).

\paragraph{Divergent Saturation Patterns.}
Beyond 28K, models exhibit divergent behaviors. Gemini and Qwen-Plus both \textit{decline}: Gemini drops from 0.876 to 0.836 ($-$4.6\%), and Qwen-Plus from 0.874 to 0.853 ($-$2.4\%). In contrast, DeepSeek continues improving from 0.868 to 0.883 (+1.7\%), demonstrating robust noise tolerance.

\paragraph{Architecture-Specific Inflection Points.}
Gemini shows minimal gains from 4K--8K before accelerating rapidly in the 12K--20K range, indicating a critical context threshold of $\sim$12K tokens. Qwen-Plus reaches 90\% of peak performance at approximately 16K tokens. DeepSeek requires 20K tokens for equivalent relative performance but continues scaling where others plateau.

\subsection{Three Scaling Behaviors}
\label{sec:scaling_behaviors}

Our analysis reveals three distinct scaling behaviors, presented in order of increasing context efficiency.

\paragraph{Cold-Start Dynamics.}
Gemini exhibits S-curve behavior: poor initial performance (0.474 at 4K) indicates limited parametric knowledge for this domain, with a critical mass requirement of $\sim$12K tokens before effective reasoning activates. Performance peaks at 28K (0.876) then declines at 32K (0.836).

\paragraph{Early Saturation.}
Qwen-Plus achieves the highest initial performance (0.689 at 4K), indicating strong parametric knowledge. However, performance peaks at 28K (0.874) and declines at 32K---extended sequences lead to attention dispersion where the model struggles to maintain focus amid noise. Optimal for short-to-medium context (4K--20K).

\paragraph{Logarithmic Growth.}
DeepSeek exhibits consistent logarithmic scaling ($R^2 \approx 0.91$), improving from 0.619 (4K) to 0.883 (32K)---the highest final performance with no decline at extended lengths. This indicates robust attention filtering suited for complex multi-document reasoning.

\paragraph{Extended Context Validation.}
Qwen-2.5-72B-Instruct extends our analysis to 128K tokens. Performance improves from 0.594 (4K) to 0.802 (32K), then plateaus at 64K (0.795) and 128K (0.805), confirming that marginal returns diminish beyond 32K. Complete results for all models are provided in Appendix~\ref{app:full_results}.

\section{Diagnostic Analysis: Metric-Level Attribution}
\label{sec:diagnostic}

The scaling curves reveal \textit{what} happens; our six-dimensional metrics diagnose \textit{why}.

\subsection{4K Context Window}
\begin{figure}[H]
    \centering
    \includegraphics[width=\linewidth]{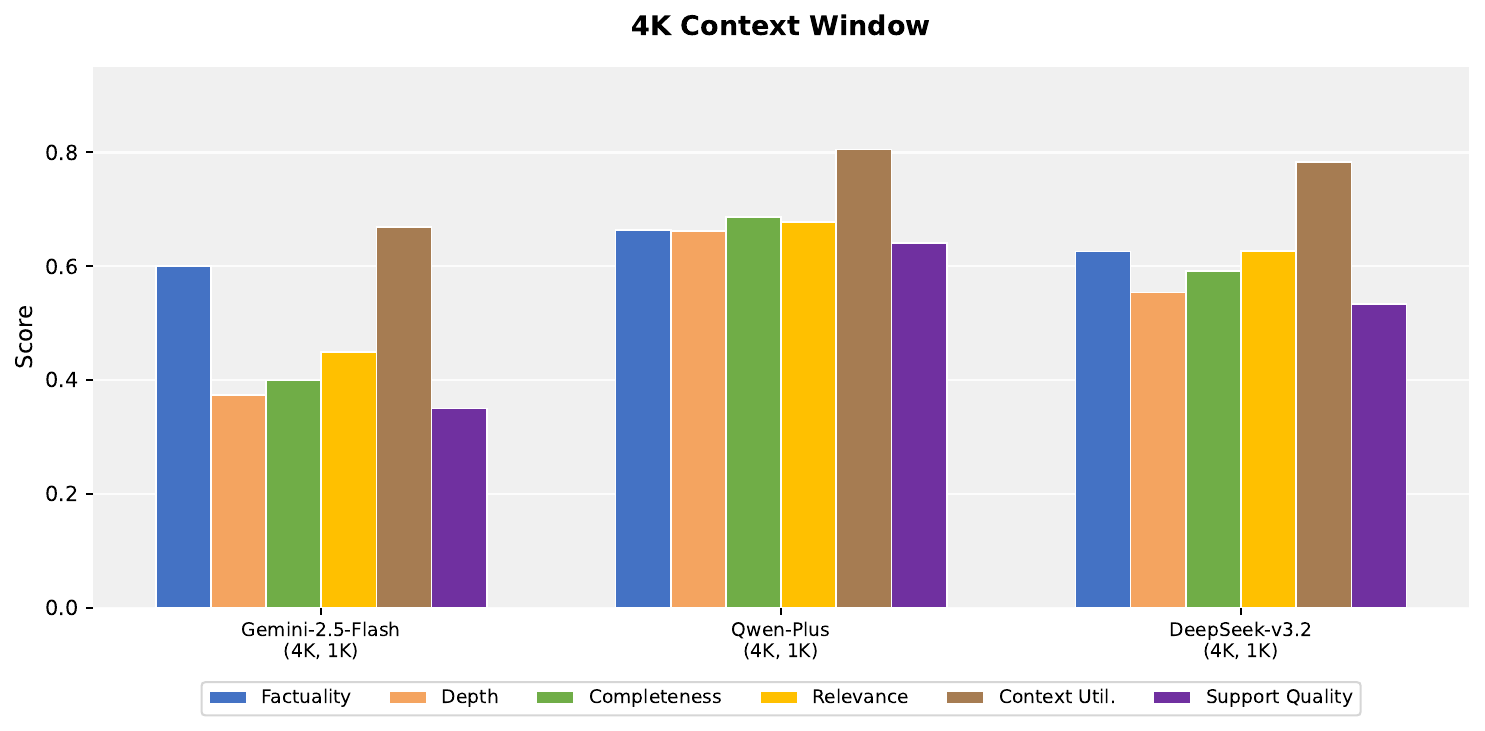}
    \caption{Metric breakdown at 4K context window---the most resource-constrained scenario.}
    \label{fig:metrics_4K}
\end{figure}

Figure~\ref{fig:metrics_4K} reveals dimension-level performance at 4K:

\textbf{Gemini} (0.474) shows catastrophically low Depth (0.374), Completeness (0.399), and Support Quality (0.350), with only Context Utilization (0.669) approaching acceptable levels---confirming inability to generate domain-specific content without extensive grounding material.

\textbf{Qwen-Plus} (0.689) leads through balanced scores: Factuality (0.663), Depth (0.662), Completeness (0.686), and notably high Context Utilization (0.805), indicating strong parametric knowledge compensating for limited retrieval.

\textbf{DeepSeek} (0.619) shows moderate performance with high Context Utilization (0.783) but lower Depth (0.554) and Support Quality (0.534), suggesting conservative evidence extraction without speculation.

\subsection{28K Context Window}
\begin{figure}[H]
    \centering
    \includegraphics[width=\linewidth]{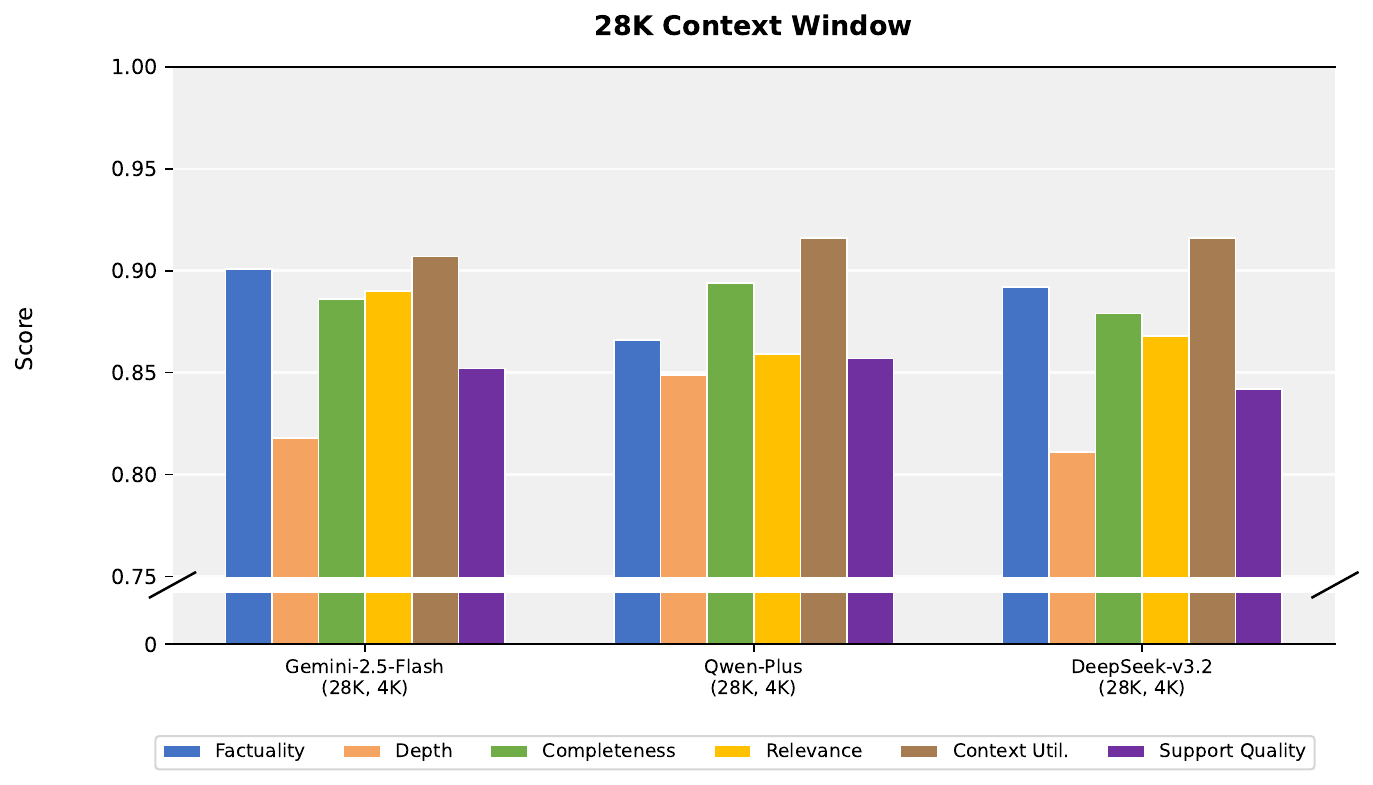}
    \caption{Metric breakdown at 28K context window (peak performance for most models).}
    \label{fig:metrics_28K}
\end{figure}

At 28K (Figure~\ref{fig:metrics_28K}), the performance landscape transforms:

\textbf{Gemini} (0.876) achieves remarkable recovery, with Factuality (0.901) and Completeness (0.886) leading all models---confirming that its short-context weakness stems from grounding dependency rather than fundamental incapability.

\textbf{Qwen-Plus} (0.874) reaches its peak with Context Utilization (0.916) and Completeness (0.894) as strongest dimensions.

\textbf{DeepSeek} (0.868) demonstrates balanced excellence with all dimensions exceeding 0.81 and no discernible weak points.

\subsection{Mechanism Attribution}

Three mechanisms explain the observed behaviors:

\begin{itemize}[leftmargin=*, itemsep=3pt]
    \item \textbf{Noise Tolerance (DeepSeek):} Monotonic improvement through 32K indicates effective attention filtering---the model benefits from additional context without information overload.
    \item \textbf{Parametric Compensation (Qwen-Plus):} Strong 4K performance reflects internal knowledge compensating for limited retrieval. The 32K decline suggests an optimal context threshold beyond which noise degrades performance.
    \item \textbf{Critical Mass Activation (Gemini):} S-curve behavior with 4K--8K stagnation followed by 12K--28K acceleration indicates a $\sim$12K token threshold for reasoning activation in this domain.
\end{itemize}

\subsection{Attention Dilution at Extreme Context (32K)}
\label{sec:attention_dilution}

\begin{figure}[H]
    \centering
    \includegraphics[width=\linewidth]{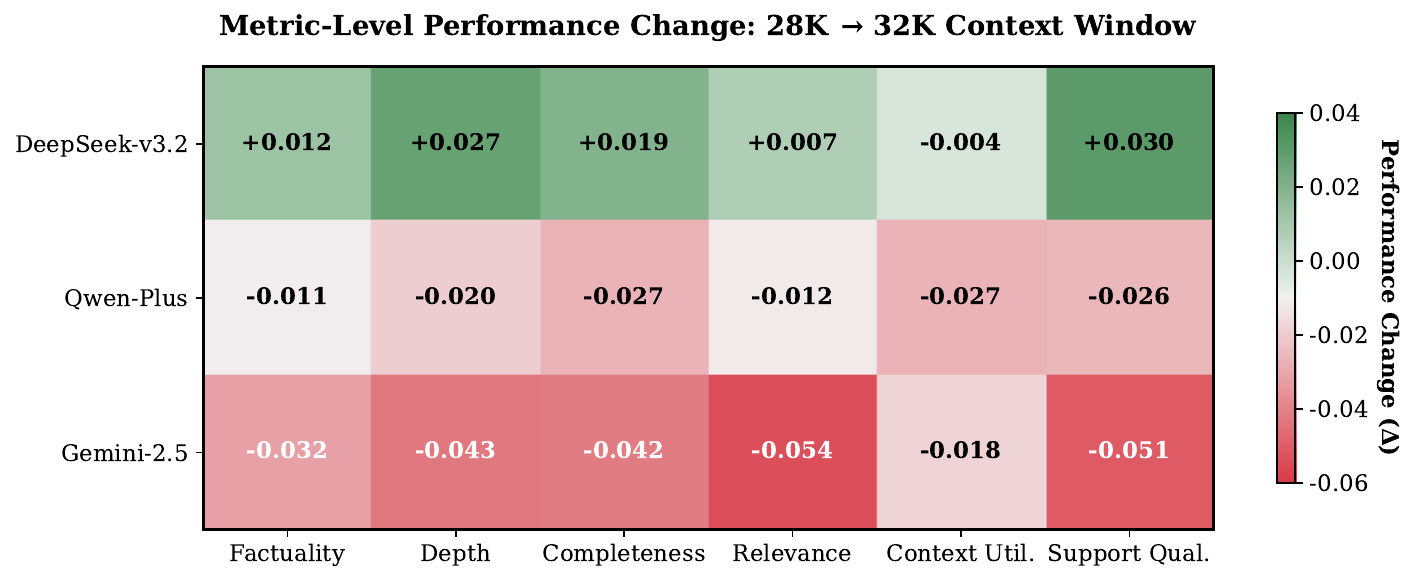}
    \caption{Metric-level performance changes from 28K to 32K context window. Green: improvement; red: decline. Gemini shows severe attention dilution with Relevance ($-$5.4\%) and Support Quality ($-$5.1\%) most affected.}
    \label{fig:attention_dilution}
\end{figure}

Figure~\ref{fig:attention_dilution} presents metric-level changes from 28K to 32K, revealing a consistent \textit{attention dilution signature} in declining models.

\paragraph{Attention Dilution Signature.}
For Gemini, the largest drops occur in \textbf{Relevance} ($-$6.1\%) and \textbf{Support Quality} ($-$6.0\%), indicating failure to identify query-relevant information. The \textbf{Depth} decline ($-$5.3\%) suggests scattered attention prevents deep analysis. Critically, \textbf{Context Utilization} shows relatively smaller decline (Gemini $-$2.0\%, Qwen-Plus $-$2.9\%), indicating these models still \textit{attempt} to leverage extended context but extract less value per token. This distinguishes attention dilution from context truncation.

\paragraph{Contrast with Noise-Tolerant Architecture.}
DeepSeek continues improving at 32K (+1.7\%), with Support Quality improvement (+3.6\%) paired with slight Context Utilization decline ($-$0.4\%)---indicating \textit{selective} utilization where the model processes more context but references only high-quality evidence.

\subsection{Metric Independence Analysis}
\label{sec:metric_independence}

To validate that our six metrics capture distinct failure modes rather than collapsing into a single quality factor, we analyze inter-metric correlations across all experimental configurations ($\sim$33 model-context combinations $\times$ 6 dimensions).

\begin{table}[H]
\centering
\small
\caption{Inter-metric Pearson correlation coefficients computed across all model-context configurations. Values below 0.70 indicate meaningful independence; values above 0.85 suggest potential redundancy.}
\label{tab:metric_correlation}
\begin{tabular}{lcccccc}
\toprule
 & \textbf{Fact.} & \textbf{Depth} & \textbf{Comp.} & \textbf{Rel.} & \textbf{Ctx.U.} & \textbf{Supp.} \\
\midrule
Factuality & 1.00 & & & & & \\
Depth & 0.96 & 1.00 & & & & \\
Completeness & 0.97 & 0.99 & 1.00 & & & \\
Relevance & 0.98 & 0.99 & 0.99 & 1.00 & & \\
Context Util. & 0.90 & 0.93 & 0.93 & 0.92 & 1.00 & \\
Support Quality & 0.98 & 0.99 & 0.99 & 0.99 & 0.91 & 1.00 \\
\bottomrule
\end{tabular}
\end{table}

Table~\ref{tab:metric_correlation} reveals high aggregate correlations ($>$0.90) across all metric pairs when computed over mean scores per model-context configuration. This is expected: as context increases, \textit{all} metrics improve together because models receive more relevant information. However, this aggregate correlation masks the diagnostic value that emerges at specific operating points.

\paragraph{Diagnostic Independence at Fixed Operating Points.}
The metrics' diagnostic utility manifests when comparing models at identical context windows, where inter-metric \textit{profiles} diverge meaningfully:

\begin{itemize}[leftmargin=*, itemsep=2pt]
    \item At 4K, Gemini's Context Utilization (0.669) is 79\% higher than its Support Quality (0.350)---a 2.3$\times$ ratio indicating the model \textit{reads} context but cannot \textit{cite} it accurately.
    \item At 28K, Gemini achieves the highest Factuality (0.901) but lower Depth (0.818), while DeepSeek shows the opposite pattern (Factuality: 0.892, but highest balanced profile). These cross-metric divergences would be invisible to a single composite score.
    \item In the case study (Section~\ref{sec:case_study}), Gemini-2.5-Flash achieves high Context Utilization (0.90) but catastrophic Factuality (0.17)---a failure mode uniquely identifiable through multi-dimensional evaluation.
\end{itemize}

\paragraph{Distinct Sensitivity to Context Scaling.}
Different metrics respond differently to context expansion: Context Utilization saturates earliest (reaching $>$0.80 by 8K for all models), while Technical Depth and Support Quality show the steepest improvement trajectories and the largest model-specific variation. The 28K-to-32K attention dilution (Section~\ref{sec:attention_dilution}) disproportionately affects Relevance and Support Quality while largely preserving Context Utilization---a diagnostic pattern only visible through multi-dimensional evaluation.

We acknowledge that the high aggregate correlations limit the metrics' discriminative power for ranking models at a single context point. The primary diagnostic value lies in \textit{profile analysis}---comparing metric patterns across models, context windows, and failure cases---rather than individual metric rankings.

\section{Cross-Framework Evaluation}
\label{sec:cross_framework}

To validate evaluation portability, we deployed our benchmark on three additional production RAG frameworks: RAGFlow, MaxKB, and Metaso.

\begin{figure}[H]
    \centering
    \includegraphics[width=\linewidth]{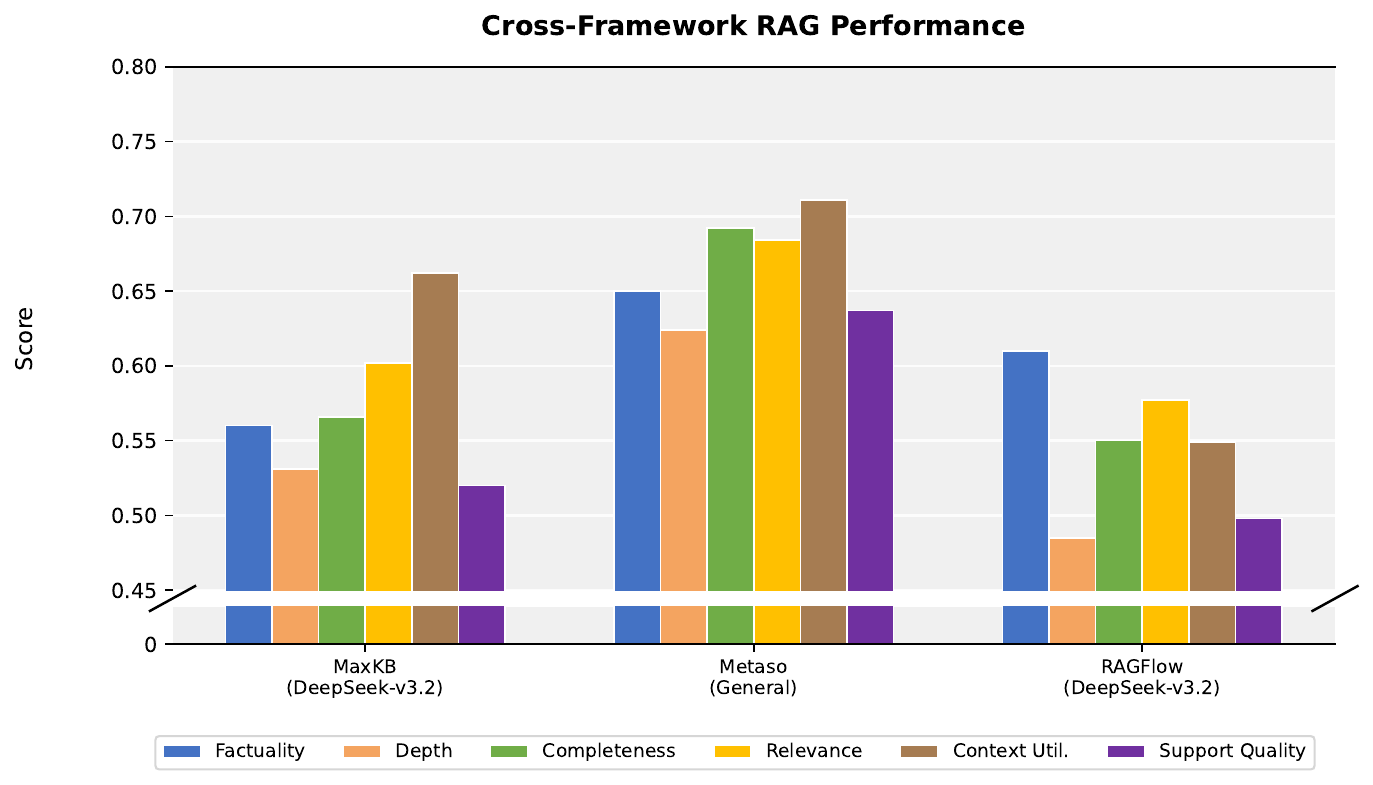}
    \caption{Performance comparison across three external RAG frameworks.}
    \label{fig:cross_framework}
\end{figure}

\begin{table}[H]
\centering
\caption{Cross-framework performance breakdown. All frameworks evaluated on the same 200-question benchmark. Retrieval strategies differ: RAGFlow uses visual document parsing with hybrid search; MaxKB uses DeepSeek-V3.2 (non-thinking mode) with chunk-based retrieval; Metaso uses a proprietary model with web-augmented retrieval.}
\label{tab:cross_framework}
\small
\begin{tabular}{lccc}
\toprule
\textbf{Metric} & \textbf{RAGFlow} & \textbf{MaxKB} & \textbf{Metaso} \\
\midrule
Factuality       & 0.610 & 0.560 & 0.650 \\
Depth            & 0.485 & 0.531 & 0.624 \\
Completeness     & 0.550 & 0.566 & 0.692 \\
Relevance        & 0.577 & 0.602 & 0.684 \\
Context Util.    & 0.549 & 0.662 & 0.711 \\
Support Quality  & 0.498 & 0.520 & 0.637 \\
\midrule
\textit{Average} & 0.545 & 0.574 & 0.666 \\
\bottomrule
\end{tabular}
\end{table}

\paragraph{Framework-Specific Failure Modes.}
Each framework exhibits distinct metric profiles reflecting different architectural choices. \textbf{Metaso} (0.67 avg) leads in Context Utilization (0.711) and Completeness (0.692), suggesting effective information synthesis. \textbf{MaxKB} (0.57 avg) shows a notable gap between Context Utilization (0.662) and Depth (0.531), indicating surface-level extraction without deep reasoning---consistent with its use of DeepSeek-V3.2 in non-thinking mode. \textbf{RAGFlow} (0.55 avg) scores lowest in Depth (0.485) and Support Quality (0.498), suggesting chunk fragmentation issues despite its advanced visual parsing capabilities.

\paragraph{Consistent Difficulty Ordering.}
Depth and Support Quality consistently score lower than Factuality across all frameworks, confirming the benchmark captures intrinsic task difficulty independent of platform architecture.

\paragraph{Deployment Portability.}
Adapting our benchmark to each new framework required minimal engineering: API integration for query submission, response parsing, and source citation extraction. The core evaluation pipeline remained unchanged, suggesting our methodology can serve as a reusable evaluation layer for heterogeneous enterprise RAG deployments.

\section{Case Study: Pulsed Atomic Layer Etching}
\label{sec:case_study}

To illustrate how multi-dimensional evaluation reveals failure modes invisible to single-score metrics, we present a representative case from technical parameter extraction.

\subsection{Task Description}

Test case MULTI\_069 requires extracting precise process parameters from patent documentation describing pulsed atomic layer etching (ALE) for ruthenium removal---a critical BEOL process in advanced semiconductor manufacturing. The fill-in-the-blank question requires five specific values: optimal bias voltage range, comparison direction, etch rate comparison, and synergy comparison relative to continuous ALE.

The ground truth specifies: \textbf{600V--1200V} bias window, \textbf{higher} bias voltages (vs.\ 60--100V for continuous ALE), \textbf{higher} etch rates (5--6~\AA/cycle vs.\ 2--3~\AA/cycle), and \textbf{higher} synergy.

\subsection{Model Responses and Error Analysis}

\begin{table}[H]
\centering
\caption{Model responses for pulsed ALE parameter extraction (MULTI\_069, 18K context).}
\label{tab:case_study_responses}
\small{
    \begin{tabular}{lccccc}
    \toprule
    \textbf{Model} & \textbf{Voltage Range} & \textbf{Direction} & \textbf{Etch Rate} & \textbf{Synergy} & \textbf{Factuality} \\
    \midrule
    Ground Truth & 600--1200V & higher & faster & increased & --- \\
    \midrule
    DeepSeek-V3.2 & 600--1200V & higher & faster & increased & \textbf{0.99} \\
    Qwen-Plus & 50--150V & lower & higher & enhanced & 0.11 \\
    Gemini-2.5-Flash & 10--50V & lower & higher & higher & 0.17 \\
    \bottomrule
    \end{tabular}
}
\end{table}

Both Qwen-Plus and Gemini exhibit \textit{process variant confusion}: the retrieved context interleaves continuous ALE (60--100V) and pulsed ALE (600--1200V) specifications. Qwen-Plus extracts values near the continuous range (50--150V), while Gemini hallucinates 10--50V. DeepSeek correctly disambiguates by reasoning about the pulsed duty cycle mechanism.

\subsection{Diagnostic Attribution via Multi-Dimensional Metrics}

\begin{figure}[H]
\centering
\includegraphics[width=0.75\linewidth]{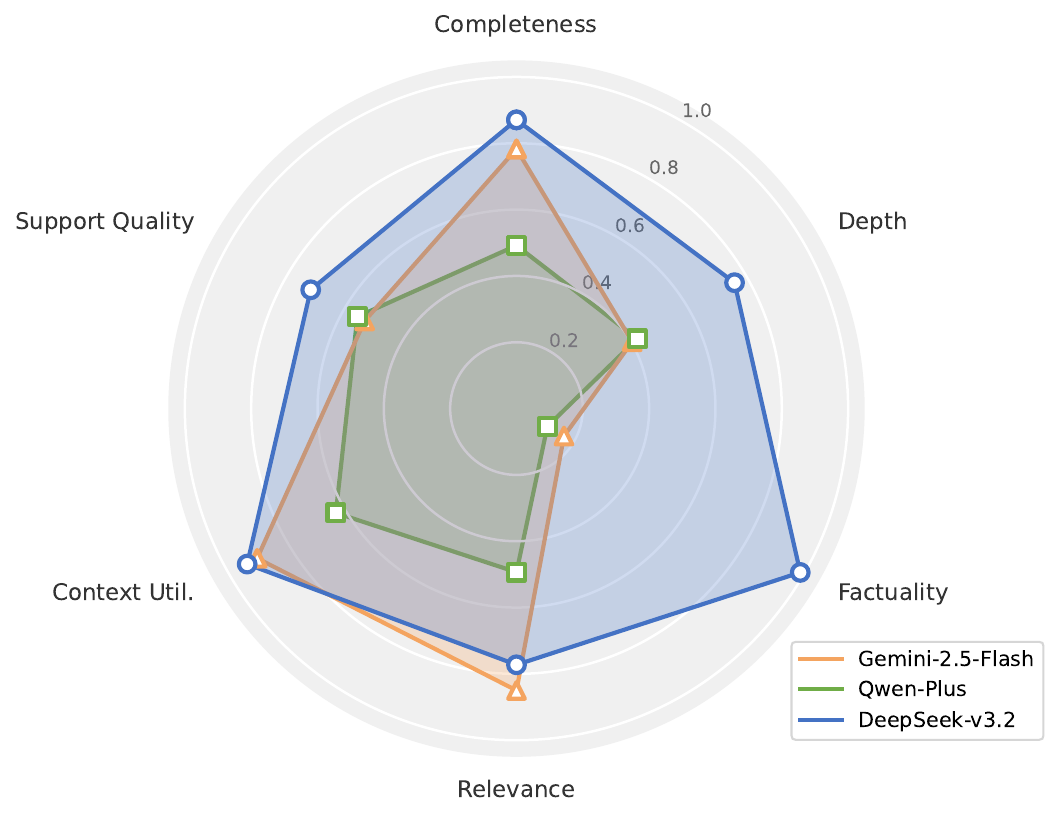}
\caption{Six-dimensional comparison for MULTI\_069 (18K context). The divergence between Context Utilization and Factuality for Gemini (0.90 vs.\ 0.17) reveals confident misattribution---a failure mode invisible to single-score evaluation.}
\label{fig:case_study_radar}
\end{figure}

The radar chart (Figure~\ref{fig:case_study_radar}) reveals a critical diagnostic pattern: Gemini achieves high Context Utilization (0.90) but catastrophic Factuality (0.17). This combination indicates the model \textit{confidently cites incorrect portions} of retrieved context---confusing continuous and pulsed ALE parameters. A single composite score would obscure this specific failure mode, which has direct implications for safety-critical semiconductor process specification.

This case exemplifies \textit{disambiguation under structural similarity}: when retrieved contexts contain multiple process variants with similar descriptive patterns but different quantitative specifications, models must parse document structure, maintain attention to qualifying terms across long spans, and cross-validate extracted values.

\section{Discussion}
\label{sec:discussion}

\subsection{Interpreting Scaling Behaviors}

We connect our three scaling behaviors to prior findings. Qwen-Plus's strong 4K performance and early saturation aligns with \citet{mallen2023not}, who found models with stronger parametric knowledge show diminishing returns from retrieval. DeepSeek's sustained 32K improvement contrasts with the ``lost in the middle'' phenomenon~\citep{liu2024lost}, suggesting effective attention distribution. Gemini's S-curve resembles multi-document QA patterns requiring sufficient context density for coherent reasoning~\citep{liu2024lost}. These interpretations remain hypotheses; definitive mechanistic attribution would require access to model internals unavailable through commercial APIs.

\subsection{Dynamic Routing Strategy}

The crossover point at $\sim$15K--16K tokens motivates context-aware model routing:

\begin{itemize}[leftmargin=*, itemsep=2pt]
    \item \textbf{Short contexts ($<$14K):} Route to Qwen-Plus for maximum efficiency.
    \item \textbf{Complex reasoning ($>$16K):} Route to DeepSeek for stable multi-document synthesis.
    \item \textbf{Batch summarization (20K--28K):} Deploy Gemini for massive context ingestion, capped at 28K.
\end{itemize}

\subsection{Limitations and Validity Threats}
\label{sec:limitations}

\paragraph{LLM-as-Judge Reliability.}
Our evaluation relies on GPT-4.1-mini as the judge model via G-Eval with structured rubrics~\citep{liu-etal-2023-g,zheng2024judging}. While the QA generation pipeline has been validated through expert evaluation (Appendix~\ref{app:expert_evaluation}), full human–judge correlation studies for the six downstream evaluation metrics—using calibrated annotator pools with inter-annotator agreement metrics—remain an important direction for future work. The structured G-Eval rubrics with separate objective/subjective scoring criteria are designed to mitigate known LLM judge biases, and the high consistency of our results across four LLMs and four RAG frameworks suggests the evaluation signal is robust.

\paragraph{Benchmark Scale.}
Our 200 curated questions (from 1,300+ generated) span three synthesis strategies across 188 topics. While sufficient for identifying significant performance differences (52,800+ metric evaluations across all configurations), larger benchmarks would enable finer-grained edge-case analysis.

\paragraph{Retrieval Configuration.}
Experiments use AnythingLLM's default retrieval (max context snippets = 4, similarity threshold = 0.25). The cross-framework evaluation (Section~\ref{sec:cross_framework}) provides implicit retrieval ablation---RAGFlow, MaxKB, and Metaso employ different retrieval strategies (visual parsing, chunk-based, web-augmented) yet our metrics consistently diagnose framework-specific failure modes. Explicit retriever ablations (e.g., TF-IDF vs.\ dense retrieval) remain future work.

\paragraph{Domain Specificity.}
Semiconductor manufacturing represents an extreme case of specialized knowledge. The generation pipeline is domain-agnostic (requiring only a corpus and knowledge base), but optimal context thresholds may require recalibration for other domains.

\paragraph{Model Selection and Reproducibility.}
We evaluate three primary models and one extended model, pinning exact version strings and dates. The rapidly evolving LLM landscape means specific findings may not transfer to newer releases, though the methodology and benchmark remain applicable.

\section{Conclusion}
\label{sec:conclusion}

We present FAB-Bench, an automated framework for evaluating RAG systems on proprietary domain knowledge. Through systematic experiments on semiconductor manufacturing documentation (150+ papers, 70+ patents), we make three contributions:

First, we establish a reproducible methodology for private-domain RAG evaluation with adaptive benchmark generation and six diagnostic metrics enabling precise failure attribution. The curated 200-question benchmark dataset is released at \url{https://github.com/FuturefabAI/FAB-Bench} and is directly applicable to other vertical domains, requiring only a domain-specific corpus and knowledge base.

Second, context-window scaling analysis (4K--32K) reveals three distinct behaviors---\textit{logarithmic growth}, \textit{early saturation}, and \textit{cold-start dynamics}---with metric-level decomposition identifying \textit{attention dilution} as the mechanism behind performance degradation at extreme context lengths.

Third, cross-framework validation on four production RAG systems confirms evaluation portability and demonstrates consistent diagnostic capability across heterogeneous architectures.

\paragraph{Future Work.}
Priority extensions include: (1)~human correlation studies validating LLM judge reliability with inter-annotator agreement metrics; (2)~explicit retriever ablations comparing sparse, dense, and hybrid retrieval strategies; (3)~per-category scaling analysis disaggregating performance across robustness, multi-hop, and generation quality question types; and (4)~extension to additional vertical domains with domain-specific knowledge bases.

\newpage
\bibliographystyle{unsrtnat}
\bibliography{references}

\newpage
\appendix

\section{Semiconductor Domain Knowledge Base}
\label{app:knowledge_base}

This appendix provides the complete listing of 431 technical terms in our semiconductor domain knowledge base, organized into seven categories with associated precision weights.

\subsection{Manufacturing Processes (60 terms, weight = 1.5)}

\textbf{Core fabrication techniques:}
wafer, lithography, etching, CVD, PVD, ALD, ion implantation, annealing, CMP, cleaning, inspection, metrology, mask, exposure, develop, strip, deposition, diffusion, oxidation, epitaxy, packaging, wire bonding, flip chip, WLP, SiP, BGA, CSP

\textbf{Advanced lithography:}
immersion lithography, multi-patterning, self-aligned, spacer, hard mask, anti-reflective coating, photoresist, reticle

\textbf{Etching techniques:}
reactive ion etching, RIE, plasma etching, dry etching, wet etching, isotropic etching, anisotropic etching

\textbf{Deposition methods:}
PECVD, LPCVD, MOCVD, MBE, electroplating, sputtering, atomic layer deposition, physical vapor deposition

\textbf{CMP and cleaning:}
chemical mechanical planarization, post-CMP cleaning, megasonic cleaning, RCA clean, HF dip

\textbf{Metrology:}
overlay metrology, critical dimension measurement, ellipsometry, optical CD, scatterometry

\subsection{Materials Science (51 terms, weight = 1.3)}

\textbf{Substrate materials:}
silicon, GaAs, GaN, SiC, InP, germanium, polysilicon, monocrystalline, amorphous

\textbf{Dielectric materials:}
high-k, low-k, SiO2, Si3N4, Al2O3, HfO2, silicon dioxide, silicon nitride, hafnium oxide, aluminum oxide, zirconium oxide, tantalum pentoxide, ultra-low-k, porous silica, carbon-doped oxide

\textbf{Metal materials:}
copper, aluminum, tungsten, titanium, tantalum, indium, silver, gold, platinum, nickel, cobalt, ruthenium, molybdenum

\textbf{Compound semiconductors:}
gallium arsenide, gallium nitride, silicon carbide, indium phosphide, aluminum nitride, zinc oxide

\textbf{2D materials:}
graphene, MoS2, transition metal dichalcogenides

\textbf{Gate stack materials:}
metal gate, work function metal, barrier layer, capping layer, etch stop layer

\subsection{Device Physics (62 terms, weight = 1.4)}

\textbf{Transistor types:}
MOSFET, FinFET, GAAFET, CMOS, BJT, JFET, IGBT, diode, transistor

\textbf{Advanced transistors:}
gate-all-around, nanosheet, nanowire, CFET, complementary FET, vertical transistor, TFET, tunnel FET, FDSOI, fully depleted SOI

\textbf{Memory devices:}
memory, DRAM, SRAM, NAND, NOR, FRAM, ReRAM, MRAM, flash memory, 3D NAND, PCM, phase change memory, RRAM, STT-MRAM, emerging memory

\textbf{Passive components:}
resistor, capacitor, inductor

\textbf{Integrated circuits:}
IC, integrated circuit, ASIC, SoC, system on chip, logic circuit, analog circuit, mixed-signal

\textbf{Sensors and MEMS:}
MEMS, sensor, actuator, image sensor, CMOS sensor, pressure sensor, accelerometer, gyroscope

\textbf{Specialty devices:}
RF, power device, optoelectronic, LED, laser diode, photodiode, solar cell, TFT, thin film transistor

\subsection{Process Nodes (65 terms, weight = 1.6)}

\textbf{Process nodes:}
14nm, 10nm, 7nm, 5nm, 3nm, 2nm, 1nm, 28nm, 22nm, 16nm, 12nm, 8nm, 6nm, 4nm, process node, technology node

\textbf{Lithography technologies:}
EUV, DUV, 193nm, 248nm, 365nm, ArF, KrF, i-line, g-line, extreme ultraviolet, deep ultraviolet, 13.5nm, high-NA EUV

\textbf{Process integration:}
FinFET process, SOI, FD-SOI, bulk, bulk silicon, strained silicon, strain engineering, SiGe, silicon germanium, HEMT, high electron mobility

\textbf{Process stages:}
FEOL, BEOL, MEOL, front end of line, back end of line, middle of line

\textbf{Advanced techniques:}
gate-first, gate-last, replacement gate, high-k metal gate, HKMG, multiple patterning, SADP, SAQP, self-aligned double patterning, self-aligned quadruple patterning, LELE, LFLE

\textbf{Interconnect:}
damascene, dual damascene, copper interconnect, via, through silicon via, TSV, backside power delivery

\subsection{Testing Methodologies (51 terms, weight = 1.2)}

\textbf{Reliability testing:}
reliability, yield, failure analysis, electrical test, functional test, burn-in, temperature cycling, thermal shock, humidity test

\textbf{Failure mechanisms:}
electromigration, hot carrier, NBTI, TDDB, negative bias temperature instability, time dependent dielectric breakdown, stress migration, void formation

\textbf{Quality control:}
SPC, statistical process control, parameter drift, defect density, process capability, Cpk, design of experiments, DOE

\textbf{Inspection techniques:}
critical dimension, CD-SEM, AFM, SEM, TEM, X-ray, atomic force microscopy, scanning electron microscopy, transmission electron microscopy, optical inspection, e-beam inspection

\textbf{Electrical characterization:}
I-V curve, C-V measurement, DLTS, deep level transient spectroscopy, Hall effect, mobility measurement, sheet resistance, four-point probe, contact resistance

\textbf{Wafer-level testing:}
probe test, wafer sort, parametric test, functional test, final test, package test

\subsection{Applications (62 terms, weight = 1.1)}

\textbf{Computing processors:}
AI chip, GPU, CPU, NPU, TPU, FPGA, ASIC, graphics processing unit, neural processing unit, tensor processing unit, application specific IC, DSP, digital signal processor, MCU, microcontroller

\textbf{Computing systems:}
edge computing, cloud computing, data center, server, supercomputer, quantum computing, neuromorphic computing, in-memory computing

\textbf{Consumer electronics:}
smartphone, tablet, laptop, wearable, smartwatch, AR glasses, VR headset

\textbf{Automotive:}
automotive electronics, ADAS, autonomous driving, LiDAR, radar, infotainment, powertrain

\textbf{Communication:}
5G, 6G, IoT, internet of things, wireless, RF transceiver, baseband, modem

\textbf{AI/ML:}
neural network, machine learning, deep learning, inference, training, transformer, large language model

\textbf{Emerging applications:}
blockchain, cryptocurrency, mining, virtual reality, VR, augmented reality, AR, metaverse, HPC, high performance computing

\subsection{Performance Parameters (80 terms, weight = 1.7)}

\textbf{Voltage parameters:}
threshold voltage, Vth, supply voltage, VDD, VSS, voltage, operating voltage, breakdown voltage, junction temperature

\textbf{Current parameters:}
leakage current, drive current, Ion, Ioff, current, saturation current, subthreshold current, gate leakage, junction leakage

\textbf{Power parameters:}
power consumption, static power, dynamic power, power dissipation, TDP, thermal design power, power density, energy efficiency

\textbf{Timing parameters:}
switching speed, delay, propagation delay, rise time, fall time, setup time, hold time, clock skew, clock frequency, access time

\textbf{Frequency parameters:}
frequency, bandwidth, cutoff frequency, fT, fmax, operating frequency, clock speed

\textbf{Resistance and capacitance:}
resistance, capacitance, Ron, on-resistance, gate capacitance, parasitic capacitance, RC delay, interconnect resistance

\textbf{Temperature parameters:}
operating temperature, junction temperature, thermal resistance, temperature coefficient

\textbf{Noise and linearity:}
noise, noise figure, SNR, signal to noise ratio, linearity, THD, total harmonic distortion

\textbf{RF parameters:}
gain, phase, S-parameters, impedance matching, insertion loss, return loss

\textbf{Performance metrics:}
DIBL, drain induced barrier lowering, subthreshold swing, SS, transconductance, gm, output conductance, GIDL, gate induced drain leakage, mobility, carrier mobility, saturation velocity

\subsection{Knowledge Base Design Rationale}

\paragraph{Coverage breadth:} The 431 terms span the full semiconductor technology stack from materials science to system applications.

\paragraph{Precision weighting:} Category weights ($w_i \in [1.1, 1.7]$) reflect technical rigor requirements, with parameters (1.7) and process nodes (1.6) weighted highest due to quantitative precision requirements.

\paragraph{Term granularity:} Includes both high-level concepts (e.g., ``transistor'') and specific implementations (e.g., ``FinFET'', ``GAAFET'', ``nanosheet'').

\paragraph{Temporal coverage:} Terms encompass mature technologies, current leading-edge, and emerging concepts to ensure benchmark relevance as technology advances.

\section{Generation System Implementation}
\label{app:generation_details}

\subsection{Precision Classification}

Documents are classified based on technical density $\rho(d)$ and weighted high-precision category presence (Eq.~\ref{eq:precision}). The disjunctive logic ensures documents containing critical parameters are classified as high-precision even with low overall density.

\subsection{Adaptive Temperature Control}

\paragraph{Base Temperature Ranges.}
Temperature ranges vary by precision level to balance accuracy with diversity:
\begin{equation}
[\tau_{\min}(p), \tau_{\max}(p)] =
\begin{cases}
[0.4, 0.8] & p = \text{high} \\
[0.5, 0.9] & p = \text{medium} \\
[0.6, 1.0] & p = \text{low}
\end{cases}
\end{equation}

\paragraph{Progressive Temperature.}
Base temperature progresses with successful generations to encourage diversity:
\begin{equation}
\tau_{\text{progress}}(k) = (\tau_{\max} - \tau_{\min}) \times \min\left(\frac{k}{20}, 0.8\right)
\end{equation}
where $k$ is the count of successfully generated questions.

\paragraph{Category-Specific Adjustments.}
Fine-grained adjustments based on dominant content category:
\begin{equation}
\Delta\tau_c(c^*) =
\begin{cases}
-0.10 & c^* = \text{parameters} \\
-0.08 & c^* = \text{processes} \\
-0.05 & c^* = \text{devices} \\
-0.03 & c^* = \text{testing} \\
\phantom{-}0.00 & c^* = \text{materials} \\
+0.02 & c^* = \text{manufacturing} \\
+0.05 & c^* = \text{applications}
\end{cases}
\end{equation}

\paragraph{Failure Recovery Boosts.}
Temperature increases after generation failures:
\begin{align}
\Delta\tau_{\text{attempt}}(a) &= \min(0.25, 0.08 \times a) \\
\Delta\tau_{\text{similarity}}(s) &= \min(0.15, 0.05 \times s)
\end{align}
where $a$ counts all failed attempts and $s$ tracks consecutive similarity failures.

\paragraph{Final Temperature Computation.}
\begin{equation}
\tau = \text{clip}\left(\tau_{\min} + \tau_{\text{progress}} + \Delta\tau_c + \Delta\tau_{\text{attempt}} + \Delta\tau_{\text{similarity}}, 0.1, 1.0\right)
\end{equation}

\paragraph{Complementary Nucleus Sampling.}
Adaptive top-$p$ maintains coherence during high-temperature exploration:
\begin{equation}
p_{\text{nucleus}}(\tau) =
\begin{cases}
0.95 & \tau \leq 0.4 \\
0.90 & 0.4 < \tau \leq 0.7 \\
0.85 & \tau > 0.7
\end{cases}
\end{equation}

\subsection{Adaptive Similarity Thresholds}

\paragraph{Base Thresholds by Precision Level.}
\begin{equation}
\theta_{\text{base}}(p) =
\begin{cases}
0.70 & p = \text{high} \\
0.75 & p = \text{medium} \\
0.80 & p = \text{low}
\end{cases}
\end{equation}

Lower thresholds for high-precision content acknowledge that questions about specialized topics necessarily share technical terminology while remaining substantively different.

\paragraph{Progressive Relaxation.}
\begin{equation}
\theta_{\text{sim}}(r) = \max(0.50, \theta_{\text{base}}(p) - 0.05 \times r)
\end{equation}

\paragraph{Weighted Similarity Computation.}
\begin{equation}
\text{sim}_{\text{weighted}}(q, q') = \text{Jaccard}(W_q, W_{q'}) + \alpha \cdot \text{Jaccard}(T_q, T_{q'})
\end{equation}
where $W$ denotes all word tokens, $T \subseteq W$ denotes technical terms, and $\alpha = 0.05$.

For corpora with $\geq 5$ existing questions, TF-IDF semantic similarity provides additional validation:
\begin{equation}
\text{sim}_{\text{semantic}}(q, q') = \cos(\vec{v}_{\text{TF-IDF}}(q), \vec{v}_{\text{TF-IDF}}(q'))
\end{equation}
Questions are rejected if $\max_{q' \in \mathcal{Q}} \text{sim}_{\text{semantic}}(q, q') > \theta_{\text{sim}} + 0.05$.

\subsection{Document Chunking}

Documents are segmented using sliding window approach: chunk size 512 tokens, overlap 128 tokens (25\%). Cross-document linking threshold: $\theta_{\text{link}} = 0.1$.

\section{Dynamic Prompt Templates}
\label{app:prompt_templates}

This section describes the core logic of our dynamic prompt engineering system.

\subsection{Prompt Architecture}

All prompts consist of four modular components that combine based on generation context:

\begin{enumerate}[leftmargin=*, itemsep=2pt]
\item \textbf{Base Template}: Defines role (AI Benchmark Scientist), injects context sections, specifies JSON output format
\item \textbf{Test-Type Module}: Scenario-specific instructions based on evaluation target
\item \textbf{Format Specification}: Question type requirements with balance tracking
\item \textbf{Diversity Directive}: Activated after repeated similarity failures
\end{enumerate}

\subsection{Test-Type Configurations}

Three test types target different RAG capabilities:

\begin{table}[H]
\centering
\caption{Test-type configurations}
\small
\begin{tabularx}{\linewidth}{lXcc}
\toprule
\textbf{Test Type} & \textbf{Instruction Focus} & \textbf{Primary} & \textbf{Secondary} \\
\midrule
Robustness & Find rare/specific facts that are easily overlooked (needle-in-haystack); construct unambiguous ground truth & Factuality & Depth, Completeness \\
\addlinespace
Multi-Hop & Identify logical chains requiring $\geq$2 sections; force cross-section synthesis & Completeness & Depth, Factuality \\
\addlinespace
Generation & Assume perfect context; require deep reasoning (analysis, comparison, causation) & Depth & Completeness, Factuality \\
\bottomrule
\end{tabularx}
\end{table}

\subsection{Question Format Balance}

We maintain target distribution of 40\% objective and 60\% subjective questions. The system tracks:
\begin{equation}
r_{\text{obj}} = \frac{n_{\text{obj}}}{n_{\text{obj}} + n_{\text{subj}}}
\end{equation}
When $r_{\text{obj}} < 0.5$, the system forces objective question generation.

\paragraph{Objective Questions (40\% target).}
Four subtypes with specific requirements:
\begin{itemize}[leftmargin=*, itemsep=2pt]
\item \textbf{Mathematical Calculation (30\%)}: Must extract ALL parameters from context; use context-specific formulas; show step-by-step with units
\item \textbf{Fill-in-Blank (25\%)}: Extract EXACT values with conditions and units as stated
\item \textbf{True/False (25\%)}: Statements about specific mechanisms requiring multi-step understanding; create both true and false statements
\item \textbf{Multiple Choice (20\%)}: All options in one line; only context provides distinguishing details
\end{itemize}

\paragraph{Subjective Questions (60\% target).}
Five archetypes rotate via $(\text{question\_count}) \mod 5$:
\begin{enumerate}[leftmargin=*, itemsep=1pt]
\item Definition/Specification: specific definitions or specification values
\item Process Explanation: process sequences or event chains
\item Causal Reasoning: reasons or purposes behind specifications
\item Comparative Analysis: compare/contrast related concepts
\item Problem Identification: potential problems, defects, or limitations
\end{enumerate}

\subsection{Diversity Enhancement}

When similarity failures exceed threshold ($s > 2$), a lightweight directive is injected:

\begin{quote}
\textit{DIVERSITY FOCUS: Previous $s$ uniqueness attempts suggest trying a new approach}\\
\textit{RECOMMENDED STYLE: Focus on ``[current archetype from rotation]''}
\end{quote}

This provides guidance to explore unexplored patterns without over-constraining generation.

\section{Expert Validation of Generated Benchmarks}
\label{app:expert_evaluation}

To assess generation quality, domain experts rated QA pairs from a multi-stage ablation study on four dimensions (1--5 scale): accuracy (factual correctness of answer), relevance (alignment with context), difficulty (cognitive challenge level), and diversity (novelty relative to other questions). Each condition generated 18 QA pairs (6 per question type: ROB, MULTI, GEN), with precision levels (high/medium/low) annotated per question based on the domain knowledge base classification.

\subsection{Ablation Results}

As shown in Table~\ref{tab:expert_ablation}, the full adaptive system with enhanced prompts (Group~E) produces the highest-quality outputs across all evaluated dimensions, with zero retry failures.

\begin{table}[H]
\centering
\caption{Expert evaluation of QA generation quality across ablation conditions (18 QA pairs per group, 1--5 scale).}
\label{tab:expert_ablation}
\small
\begin{tabular}{llcccccc}
\toprule
\textbf{Group} & \textbf{Configuration} & \textbf{Acc.} & \textbf{Rel.} & \textbf{Diff.} & \textbf{Div.} & \textbf{Retries} & \textbf{Time} \\
\midrule
A (Baseline)           & Fixed $\tau=0.5$, Fixed $\theta=0.65$     & 4.67 & 4.72 & 4.00 & 3.67 & 15 & 18m 33s \\
B (+Adaptive $\tau$)   & Adaptive $\tau$, Fixed $\theta=0.65$  & 4.89 & 4.83 & 4.17 & 3.39 & 16 & 30m 37s \\
C (+Adaptive $\theta$) & Fixed $\tau=0.5$, Adaptive $\theta$  & 4.61 & 4.61 & 3.61 & 3.28 & 47 & 36m 55s \\
D (Adaptive $\tau+\theta$) & Adaptive $\tau$, Adaptive $\theta$ & 4.33 & 4.78 & 3.28 & 3.17 & 12 & 32m 19s \\
E (Full System)        & Adaptive $\tau+\theta$ + enhanced prompts & \textbf{5.00} & \textbf{5.00} & \textbf{4.60} & \textbf{4.53} & \textbf{0} & 25m 15s \\
\bottomrule
\end{tabular}
\end{table}

Notably, adaptive temperature and threshold alone (Group~D) do not outperform the baseline in all dimensions---Difficulty (3.28 vs.\ 4.00) and Diversity (3.17 vs.\ 3.67) decrease, suggesting that parameter adaptation without prompt enhancement leads to conservative generation. The full system (Group~E) combines adaptive parameters with enhanced prompt templates (Appendix~\ref{app:prompt_templates}), achieving consistent improvements across all dimensions.

\subsection{Per-Question-Type Quality Analysis}

Table~\ref{tab:expert_by_type} breaks down expert ratings by question type, revealing systematic quality patterns across synthesis strategies.

\begin{table}[H]
\centering
\caption{Expert-rated quality by question type (averaged across all ablation groups).}
\label{tab:expert_by_type}
\small
\begin{tabular}{lcccc}
\toprule
\textbf{Type} & \textbf{Accuracy} & \textbf{Relevance} & \textbf{Difficulty} & \textbf{Diversity} \\
\midrule
Robustness (ROB) & 4.58 & 4.62 & 2.38 & 2.92 \\
Multi-Hop (MULTI) & 4.58 & 4.83 & 4.12 & 3.33 \\
Generation (GEN) & 4.71 & 4.75 & 4.79 & 3.88 \\
\bottomrule
\end{tabular}
\end{table}

The difficulty gradient (ROB: 2.38 $<$ MULTI: 4.12 $<$ GEN: 4.79) confirms the three synthesis strategies produce appropriately graded complexity: needle-in-haystack questions are factual and straightforward, cross-document multi-hop questions require moderate reasoning, and generation quality questions demand deep analytical reasoning. Accuracy remains high ($>$4.5) across all types, indicating factual correctness is maintained regardless of difficulty level.

\section{Evaluation Rubrics}
\label{app:rubrics}

All metrics use GPT-4.1-mini via DeepEval's G-Eval with chain-of-thought reasoning, scoring 0--10 (internally normalized to 0--1). Rubrics adapt between objective and subjective questions.

\subsection{Completeness}

\begin{table}[H]
\centering
\caption{Completeness rubric}
\small
\begin{tabularx}{\linewidth}{>{\centering\arraybackslash}p{0.8cm}XX}
\toprule
\textbf{Score} & \textbf{Objective} & \textbf{Subjective} \\
\midrule
9--10 & Correct answer with complete explanation and clear reasoning & Addresses all key aspects thoroughly \\
7--8  & Correct answer with good explanation, minor gaps & Covers all major aspects with good detail \\
5--6  & Answer with basic explanation, some missing details & Addresses most important aspects reasonably \\
3--4  & Partial answer or incomplete explanation & Covers key aspects with varying depth \\
1--2  & Minimal answer, lacks proper explanation & Addresses some but misses important points \\
0     & No clear answer or off-topic & Fails to address key aspects \\
\bottomrule
\end{tabularx}
\end{table}

\subsection{Technical Depth}

\begin{table}[H]
\centering
\caption{Technical depth rubric}
\small
\begin{tabularx}{\linewidth}{>{\centering\arraybackslash}p{0.8cm}XX}
\toprule
\textbf{Score} & \textbf{Objective} & \textbf{Subjective} \\
\midrule
9--10 & Sophisticated understanding with detailed methodology & Sophisticated understanding with nuanced analysis \\
7--8  & Strong analytical thinking with clear solution steps & Strong analytical thinking with good conceptual depth \\
5--6  & Reasonable technical detail with some analysis & Reasonable analysis with adequate technical detail \\
3--4  & Basic approach but relatively shallow analysis & Surface-level analysis with limited sophistication \\
1--2  & Minimal analytical content, superficial approach & Superficial treatment with minimal content \\
0     & No meaningful analysis demonstrated & No meaningful analysis or understanding \\
\bottomrule
\end{tabularx}
\end{table}

\subsection{Factuality}

\begin{table}[H]
\centering
\caption{Factuality rubric}
\small
\begin{tabularx}{\linewidth}{>{\centering\arraybackslash}p{0.8cm}XX}
\toprule
\textbf{Score} & \textbf{With Retrieval Context} & \textbf{Without Context} \\
\midrule
9--10 & All claims match context exactly, no contradictions & Claims match expected output, no contradictions \\
7--8  & Minor imprecision in non-critical details & Most claims consistent, no significant errors \\
5--6  & Generally accurate with minor factual gaps & Some inconsistencies but generally reasonable \\
3--4  & Noticeable inaccuracies but core info correct & Notable contradictions or implausible info \\
1--2  & Several factual errors or contradictions & Major contradictions or obviously incorrect \\
0     & Major errors or predominantly false information & --- \\
\bottomrule
\end{tabularx}
\end{table}

\subsection{Relevance}

\begin{table}[H]
\centering
\caption{Relevance rubric}
\small
\begin{tabularx}{\linewidth}{>{\centering\arraybackslash}p{0.8cm}XX}
\toprule
\textbf{Score} & \textbf{Objective} & \textbf{Subjective} \\
\midrule
9--10 & Directly addresses specific question format & Every sentence directly addresses the question \\
7--8  & Stays focused with minimal deviation & Strongly focused with minimal deviation \\
5--6  & Generally addresses question, some unnecessary details & Mostly focused with minor tangential elements \\
3--4  & Partially addresses with irrelevant information & Generally on-topic but includes unnecessary info \\
1--2  & Somewhat related with significant drift & Partially addresses with significant drift \\
0     & Fails to address specific requirements & Minimal connection to actual question \\
\bottomrule
\end{tabularx}
\end{table}

\subsection{Context Utilization}

\begin{table}[H]
\centering
\caption{Context utilization rubric}
\small
\begin{tabularx}{\linewidth}{>{\centering\arraybackslash}p{0.8cm}XX}
\toprule
\textbf{Score} & \textbf{With Retrieval Context} & \textbf{Manual Setup} \\
\midrule
9--10 & Seamlessly weaves multiple sources together & Sophisticated domain expertise with advanced terminology \\
7--8  & Effectively uses most relevant context & Clear evidence of domain-specific knowledge \\
5--6  & Incorporates key context adequately & Some evidence of specialized knowledge \\
3--4  & Uses some context, integration could improve & Minimal evidence of specialized knowledge \\
1--2  & Minimal effective use of available context & Generic response with limited domain content \\
0     & Completely ignores or contradicts context & No domain-specific content \\
\bottomrule
\end{tabularx}
\end{table}

\subsection{Support Quality}

\begin{table}[H]
\centering
\caption{Support quality rubric}
\small
\begin{tabularx}{\linewidth}{>{\centering\arraybackslash}p{0.8cm}XX}
\toprule
\textbf{Score} & \textbf{Objective} & \textbf{Subjective} \\
\midrule
9--10 & Exceptional evidence with specific calculations/formulas & Exceptional evidence with specific, concrete details \\
7--8  & Strong evidence with good specificity & Strong evidence with good specificity \\
5--6  & Adequate details with reasonable explanation & Adequate details with reasonable examples \\
3--4  & Some details but could be more specific & Some details but could be more specific \\
1--2  & Minimal evidence or poorly explained & Minimal evidence or poorly chosen examples \\
0     & Lacks evidence or contains misleading explanations & Lacks evidence or contains misleading examples \\
\bottomrule
\end{tabularx}
\end{table}

\section{Evaluation Platform Details}
\label{app:platform_details}

\subsection{Framework Capabilities}

\begin{table}[H]
\centering
\small
\caption{RAG framework integration capabilities.}
\begin{tabular}{lcccc}
\toprule
\textbf{Framework} & \textbf{API Upload} & \textbf{Sources Exposed} & \textbf{Model Config} & \textbf{Setup} \\
\midrule
AnythingLLM & \checkmark & \checkmark & \checkmark & Recommended \\
RAGFlow & \checkmark & \checkmark & \checkmark & Recommended \\
MaxKB & -- & -- & \checkmark & Required \\
Metaso & \checkmark & Partial & -- & Required \\
\bottomrule
\end{tabular}
\end{table}

\paragraph{Key Differences.}
\textbf{AnythingLLM} provides full retrieval transparency through its sources field, returning document chunks with relevance scores.

\textbf{RAGFlow} uses an OpenAI-compatible API endpoint, returning responses in standard format. Source attribution requires inference from response content.

\textbf{MaxKB} does not expose retrieval context through its API---sources are visible only in the web interface. Our adapter implements heuristic detection using domain-specific indicators (e.g., ``according to'', ``standard specifies'', ``SEMI'') and professional terminology presence.

\textbf{Metaso} returns structured references but uses a proprietary model that cannot be reconfigured, limiting participation to cross-framework evaluation.

\subsection{Adaptive Evaluation Strategy}

The evaluation layer adapts metric computation based on framework capabilities. For frameworks without source exposure, Context Utilization switches from direct source comparison to heuristic inference based on response characteristics, ensuring fair evaluation across frameworks with different transparency levels.

\subsection{Dual-Mode Assessment}

\begin{itemize}[leftmargin=*, itemsep=2pt]
\item \textbf{Mode A (with\_kb)}: Standard RAG query through framework's native retrieval pipeline
\item \textbf{Mode B (without\_kb)}: Gold context injected into prompt, bypassing retrieval to isolate generation capabilities
\end{itemize}

Cross-mode score differences ($\Delta = \text{Mode B} - \text{Mode A}$) enable failure attribution: large positive $\Delta$ indicates retrieval failures; consistently low scores in both modes indicate generation-stage weaknesses.

\section{Complete Experimental Results}
\label{app:full_results}

\begin{table}[H]
\centering
\caption{DeepSeek-v3.2-Exp: complete results across all context configurations.}
\small
\begin{tabular}{lcccccc|c}
\toprule
\textbf{Config} & \textbf{Fact.} & \textbf{Depth} & \textbf{Comp.} & \textbf{Rel.} & \textbf{Ctx.U.} & \textbf{Supp.} & \textbf{Avg.} \\
\midrule
(4K, 1K)  & 0.626 & 0.554 & 0.592 & 0.627 & 0.783 & 0.534 & 0.619 \\
(8K, 2K)  & 0.632 & 0.602 & 0.640 & 0.661 & 0.804 & 0.594 & 0.656 \\
(10K, 4K) & 0.663 & 0.606 & 0.651 & 0.669 & 0.814 & 0.609 & 0.669 \\
(12K, 4K) & 0.690 & 0.640 & 0.688 & 0.701 & 0.824 & 0.644 & 0.698 \\
(14K, 4K) & 0.744 & 0.693 & 0.752 & 0.740 & 0.845 & 0.709 & 0.747 \\
(16K, 4K) & 0.792 & 0.718 & 0.783 & 0.781 & 0.868 & 0.735 & 0.780 \\
(18K, 4K) & 0.837 & 0.755 & 0.820 & 0.813 & 0.884 & 0.776 & 0.814 \\
(20K, 4K) & 0.865 & 0.796 & 0.860 & 0.850 & 0.901 & 0.823 & 0.849 \\
(24K, 4K) & 0.868 & 0.812 & 0.880 & 0.857 & 0.915 & 0.839 & 0.862 \\
(28K, 4K) & 0.892 & 0.811 & 0.879 & 0.868 & 0.916 & 0.842 & 0.868 \\
(32K, 4K) & \textbf{0.904} & \textbf{0.838} & \textbf{0.898} & \textbf{0.875} & 0.912 & \textbf{0.872} & \textbf{0.883} \\
\bottomrule
\end{tabular}
\end{table}

\begin{table}[H]
\centering
\caption{Qwen-Plus: complete results across all context configurations.}
\small
\begin{tabular}{lcccccc|c}
\toprule
\textbf{Config} & \textbf{Fact.} & \textbf{Depth} & \textbf{Comp.} & \textbf{Rel.} & \textbf{Ctx.U.} & \textbf{Supp.} & \textbf{Avg.} \\
\midrule
(4K, 1K)  & 0.663 & 0.662 & 0.686 & 0.677 & 0.805 & 0.641 & 0.689 \\
(8K, 2K)  & 0.665 & 0.710 & 0.724 & 0.704 & 0.801 & 0.696 & 0.717 \\
(10K, 4K) & 0.680 & 0.736 & 0.764 & 0.736 & 0.828 & 0.724 & 0.745 \\
(12K, 4K) & 0.697 & 0.749 & 0.765 & 0.722 & 0.823 & 0.747 & 0.751 \\
(14K, 4K) & 0.723 & 0.772 & 0.804 & 0.760 & 0.834 & 0.767 & 0.777 \\
(16K, 4K) & 0.750 & 0.789 & 0.809 & 0.768 & 0.848 & 0.770 & 0.789 \\
(18K, 4K) & 0.732 & 0.783 & 0.813 & 0.774 & 0.844 & 0.783 & 0.788 \\
(20K, 4K) & 0.778 & 0.810 & 0.830 & 0.798 & 0.852 & 0.801 & 0.812 \\
(24K, 4K) & 0.833 & 0.829 & 0.861 & 0.827 & 0.882 & 0.835 & 0.845 \\
(28K, 4K) & \textbf{0.866} & \textbf{0.849} & \textbf{0.894} & \textbf{0.859} & \textbf{0.916} & \textbf{0.857} & \textbf{0.874} \\
(32K, 4K) & 0.855 & 0.829 & 0.867 & 0.847 & 0.889 & 0.831 & 0.853 \\
\bottomrule
\end{tabular}
\end{table}

\begin{table}[H]
\centering
\caption{Gemini-2.5-Flash: complete results across all context configurations.}
\small
\begin{tabular}{lcccccc|c}
\toprule
\textbf{Config} & \textbf{Fact.} & \textbf{Depth} & \textbf{Comp.} & \textbf{Rel.} & \textbf{Ctx.U.} & \textbf{Supp.} & \textbf{Avg.} \\
\midrule
(4K, 1K)  & 0.600 & 0.374 & 0.399 & 0.449 & 0.669 & 0.350 & 0.474 \\
(8K, 2K)  & 0.643 & 0.536 & 0.563 & 0.583 & 0.757 & 0.523 & 0.601 \\
(10K, 2K) & 0.666 & 0.550 & 0.595 & 0.622 & 0.758 & 0.622 & 0.636 \\
(12K, 4K) & 0.701 & 0.641 & 0.692 & 0.708 & 0.819 & 0.651 & 0.702 \\
(14K, 4K) & 0.743 & 0.655 & 0.709 & 0.726 & 0.839 & 0.656 & 0.721 \\
(16K, 4K) & 0.760 & 0.669 & 0.733 & 0.731 & 0.816 & 0.676 & 0.731 \\
(18K, 4K) & 0.776 & 0.710 & 0.771 & 0.784 & 0.847 & 0.728 & 0.769 \\
(20K, 4K) & 0.833 & 0.738 & 0.804 & 0.808 & 0.878 & 0.752 & 0.802 \\
(24K, 4K) & 0.854 & 0.765 & 0.827 & 0.837 & 0.882 & 0.790 & 0.826 \\
(28K, 4K) & \textbf{0.901} & \textbf{0.818} & \textbf{0.886} & \textbf{0.890} & \textbf{0.907} & \textbf{0.852} & \textbf{0.876} \\
(32K, 4K) & 0.869 & 0.775 & 0.844 & 0.836 & 0.889 & 0.801 & 0.836 \\
\bottomrule
\end{tabular}
\end{table}

\begin{table}[H]
\centering
\caption{Qwen-2.5-72B-Instruct: results with extended context windows up to 128K.}
\small
\begin{tabular}{lcccccc|c}
\toprule
\textbf{Config} & \textbf{Fact.} & \textbf{Depth} & \textbf{Comp.} & \textbf{Rel.} & \textbf{Ctx.U.} & \textbf{Supp.} & \textbf{Avg.} \\
\midrule
(4K, 1K)   & 0.511 & 0.568 & 0.603 & 0.589 & 0.758 & 0.536 & 0.594 \\
(8K, 2K)   & 0.544 & 0.590 & 0.629 & 0.605 & 0.785 & 0.560 & 0.619 \\
(16K, 4K)  & 0.643 & 0.661 & 0.713 & 0.692 & 0.798 & 0.630 & 0.690 \\
(32K, 4K)  & 0.794 & 0.762 & 0.827 & 0.812 & 0.852 & 0.764 & 0.802 \\
(64K, 6K)  & 0.801 & 0.759 & 0.820 & 0.796 & 0.856 & 0.738 & 0.795 \\
(128K, 8K) & 0.797 & 0.766 & 0.829 & 0.812 & \textbf{0.863} & 0.762 & \textbf{0.805} \\
\bottomrule
\end{tabular}
\end{table}

\begin{table}[H]
\centering
\caption{Cross-model comparison at key context configurations.}
\label{tab:summary_comparison}
\small
\begin{tabular}{lcccc}
\toprule
\textbf{Model} & \textbf{4K} & \textbf{16K} & \textbf{28K} & \textbf{32K} \\
\midrule
DeepSeek-v3.2-Exp & 0.619 & 0.780 & 0.868 & \textbf{0.883} \\
Qwen-Plus & \textbf{0.689} & \textbf{0.789} & \textbf{0.874} & 0.853 \\
Gemini-2.5-Flash & 0.474 & 0.731 & 0.876 & 0.836 \\
Qwen-2.5-72B & 0.594 & 0.690 & --- & 0.802 \\
\bottomrule
\end{tabular}
\end{table}

\end{document}